\documentclass[twoside,11pt]{article}
\usepackage{haldefs}
\usepackage{jmlr2e}
\usepackage{url}
\usepackage{algorithm}
\usepackage{algorithmic}

\jmlrheading{1}{2000}{1-48}{4/00}{10/00}{Hal Daum\'e III and Daniel Marcu}

\ShortHeadings{A Bayesian Model for Supervised Clustering}{Daum\'e III and Marcu}
\firstpageno{1}

\begin{document}

\title{A Bayesian Model for Supervised Clustering\\with the Dirichlet Process Prior}

\author{\name Hal Daum\'e III \email hdaume@isi.edu\\
        \name Daniel Marcu \email marcu@isi.edu\\
        \addr Information Sciences Institute \\
              University of Southern California \\
              4676 Admiralty Way, Suite 1001 \\
              Marina del Rey, CA 90292, USA }

\newcommand{\nth}{^{(n)}}
\newcommand{\noth}{^{(n+1)}}

\newcommand{\hide}[1]{}

\newcommand{\system}[1]{\textsc{#1}}
\newcommand{\bsf}[1]{{\bf \textsf{#1}}}
\editor{William Cohen}

\maketitle

\begin{abstract}
We develop a Bayesian framework for tackling the supervised clustering
problem, the generic problem encountered in tasks such as reference
matching, coreference resolution, identity uncertainty and record
linkage.  Our clustering model is based on the Dirichlet process
prior, which enables us to define distributions over the countably
infinite sets that naturally arise in this problem.  We add
\emph{supervision} to our model by positing the existence of a set of
unobserved random variables (we call these ``reference types'') that
are generic across all clusters.  Inference in our framework, which
requires integrating over infinitely many parameters, is solved using
Markov chain Monte Carlo techniques.  We present algorithms for both
conjugate and non-conjugate priors.  We present a simple---but
general---parameterization of our model based on a Gaussian
assumption.  We evaluate this model on one artificial task and three
real-world tasks, comparing it against both unsupervised and
state-of-the-art supervised algorithms.  Our results show that our
model is able to outperform other models across a variety of tasks and
performance metrics.
\end{abstract}


\section{Introduction}

Supervised clustering is the general characterization of a problem
that occurs frequently in strikingly different communities.  Like
standard clustering, the problem involves breaking a finite set $X
\subseteq \cX$ into a $K$-way partition $B_1, \dots, B_K$ (with $K$
unknown).  The distinction between supervised clustering and standard
clustering is that in the supervised form we are given training
examples.  These training examples enable a learning algorithm to
determine what aspects of $X$ are relevant to creating an appropriate
clustering.  The $N$ training examples $(X\nth, \{B_k\}_{k=1\dots
K\nth}\nth)$ are subsets of $\cX$ paired with their correct
partitioning.  In the end, the supervised clustering task is a
prediction problem: a new $X\noth \subseteq \cX$ is presented and a
system must produce a partition of it.

The supervised clustering problem goes under many names, depending on
the goals of the interested community.  In the relational learning
community, it is typically referred to as \emph{identity uncertainty}
and the primary task is to augment a reasoning system so that it does
not implicitly (or even explicitly) assume that there is a one-to-one
correspondence between elements in an knowledge base and entities in
the real world \citep{cohen02match,pasula03iu}.  In the database
community, the task arises in the context of merging databases with
overlapping fields, and is known as \emph{record linkage}
\citep{monge97database,doan04ontology}.  In information extraction,
particularly in the context of extracting citations from scholarly
publications, the task is to identify which citations are to the same
publication.  Here, the task is known as \emph{reference matching}
\citep{mccallum00matching}.  In natural language processing, the
problem arises in the context of \emph{coreference resolution},
wherein one wishes to identify which entities mentioned in a document
are the same person (or organization) in real life
\citep{soon01machine,ng-cardie-02,mccallum04coref}.  In the machine
learning community, it has additionally been referred to as
\emph{learning under equivalence constraints} \citep{bar-hillel03eq}
and \emph{learning from cluster examples} \citep{kamishima03lce}.

In this paper, we propose a generative model for solving the
supervised clustering problem.  Our model takes advantage of the
\emph{Dirichlet process prior}, which is a non-parametric Bayesian
prior over discrete distributions.  This prior plays two crucial
roles: first, it allows us to estimate the number of clusters $K$ in a
principled manner; second, it allows us to control the complexity of
the solutions that are learned.  We present inference methods for our
model based on Markov chain Monte Carlo methods.  We compare our model
against other methods on large, real-world data sets, where we show
that it is able to outperform most other systems according to several
metrics of performance.

The remainder of this paper is structured as follows.  In
Section~\ref{sec:prior}, we describe prior efforts to tackle the
supervised clustering problem.  In Section~\ref{sec:model}, we develop
our framework for this problem, starting from very basic assumptions
about the task.  We follow this discussion with a general scheme for
inference in this framework (Section~\ref{sec:inference}).  Next, in
Section~\ref{sec:param}, we present three generic parameterizations of
our framework and describe the appropriate adaptation of the inference
scheme to these parameterizations.  We then discuss performance
metrics for the supervised clustering problem in
Section~\ref{sec:metrics} and present experimental results of our
models' performance on artificial and real-world problems in
Section~\ref{sec:results}.  We conclude in
Section~\ref{sec:discussion} with a discussion of the advantages and
disadvantages of our model, our generic parameterization, and our
learning techniques.

\section{Prior Work} \label{sec:prior}

The most common technique for solving supervised clustering is by
mapping it to binary classification.  For a given input set, a binary
classifier is trained on all pairs of inputs, eliciting a positive
output if the two elements belong in the same cluster and a negative
output otherwise.  When applied to test data, however, such a
classifier will not necessarily produce a valid equivalence relation
(i.e., it might say $x = y$ and $y = z$ but $x \neq z$); to solve this
problem, the outputs of the binary classifier are fed into a
clustering algorithm.  Among others, \citet{cohen02match} present an
agglomerative clustering algorithm in the task of record linkage;
\citet{bar-hillel03eq} present a similar, but more complex algorithm
that is provably optimal whenever the binary classifier is
sufficiently good.\footnote{Unfortunately, the ``sufficiently good
requirement'' of \citet{bar-hillel03eq} is often unattainable: it
states that the classifier must achieve an error rate of at most
$R^2/6$, where $R$ is the ratio of the size of the smallest class to
the total number of points.  In many real world problems, the size of
the smallest class is $1$, and the number of points is quite large,
meaning that only a perfect classifier will achieve the required
accuracy.}  

The binary classification plus clustering approach is attractive
primarily because both of these problems have individually received
much attention; thus, good algorithms are known to solve them.  The
primary disadvantages of these approaches are the largely ad-hoc
connection between the classifier and the clustering algorithm, the
necessity of training over $\O(n^2)$ data points, and the potential
difficulty of performing unbiased cross-validation to estimate
hyperparameters.  The first issue, the ad-hoc connection, makes it
difficult to make state precise statements about performance.  The
second can cause computational problems for expensive classifiers
(such as SVMs) and invalidates the i.i.d. assumption that is necessary
for many generalization bounds.\footnote{For instance, the pairs
$(x_1,x_2)$ and $(x_3,x_4)$ can be seen as being drawn i.i.d. from a
joint pair distribution, but the pairs $(x_1,x_2)$, $(x_2,x_3)$ cannot
possibly be i.i.d.}  The final issue, regarding cross-validation, has
to do with the fact that the classification plus clustering approach
is based on pipelining two independent systems (see
Section~\ref{sec:results:systems} for how the cross-validation is done
in our comparative model).

In addition to the classification plus clustering approach, there have
been several attempts to solve the supervised clustering problem
directly.  Some researchers have posed the problem in the framework of
learning a distance metric, for which, eg., convex optimization
methods can be employed
\citep{bar-hillel03dist,xing03metric,basu03semisupervised}.  Using a
learned distance metric, one is able to use a standard clustering
algorithm for doing the final predictions.  These methods effectively
solve all of the problems associated with the classification plus
clustering approach.  The only drawback to these approaches is that
they assume Euclidean data and learn a Mahalanobis distance metric.
It is often unclear how to extend this assumption to a more general
space or a more general notion of similarity.

Two other recent techniques have been proposed for directly solving
the supervised clustering problem, and are not phrased in terms of
learning a Mahalanobis distance.  The first, due to
\citet{mccallum04coref}, is based on conditional random fields.  In
this model, a fully connected graph is created, where nodes are
elements in a data set.  Feature functions are defined over the edges
(corresponding to pairs of input elements), and weights are learned to
maximize the conditional likelihood of the data.  In order to ensure
that the model never predicts intransitive solutions, clique
potentials of $-\infty$ are inserted for any solution that is
intransitive.  Exact inference in this model is intractable (as in
most supervised clustering models), and they employ a simple
perceptron-style update scheme, which they show to be quite effective
on this task.  The perceptron requires that the most likely clustering
be found for a given set of weights, which is NP-complete by reduction
to graph partitioning; \citet{mccallum04coref} employ a standard
approximation algorithm for performing this operation.  This technique
appears promising, largely because it can incorporate arbitrary
feature functions.  The only potential drawback seems to be that two
approximations are used: the perceptron approximation to the CRF
likelihood\footnote{It could be argued that the perceptron
``approximation'' is actually superior to the CRF, since it optimizes
something closer to ``accuracy'' than the log-loss optimized by the
CRF.} and an approximate graph partitioning algorithm for performing
the clustering.

The other direct solution to the supervised clustering problem, due to
\citet{finley05sc}, is based on the SVMs for Interdependent and
Structured Outputs technique \citep{tsochantaridis04svmiso}.  In this
model, a particular clustering method, \emph{correlation clustering},
is held fixed, and weights are optimized to minimize the regularized
empirical loss of the training data with respect to this clustering
function.  The choice of correlation clustering is not accidental: it
decomposes over pairs.  The advantage of this model over the model of
\citet{mccallum04coref} is primarily due to the fact that the SVM
model can optimize more complex (and appropriate) loss functions than
can the CRF approach.  However, like the CRF approach, the SVMISO
approach must resort to approximation methods for finding solutions
during learning.

In comparison to other models that have been proposed, ours most
closely resembles the (non-Bayesian) generative model proposed by
\citet{pasula03iu}.  This model formulates the identity
uncertainty/citation matching problem in a generative framework, based
on a complex generative model under which inference is intractable.
They resort to an Markov chain Monte Carlo inference scheme for
identifying clusters, where a uniform prior is placed on the number of
clusters.  Their framework learns the model parameters through an MCMC
sampling procedure, though no learning is done with respect to the
prior on the number of clusters.  The work we present in this paper
can be seen as a method for extending their approach in two ways:
first, we directly model the number of output clusters; second, we
provide an intuitive, effective procedure for accounting for the
multiple aspects of similarity between different instances.  As we
discuss in Section~\ref{sec:discussion}, the hybridization of their
model and the one we propose could lead to a more effective system
than either alone.  (Indeed, between the time of submission of this
paper and its final acceptance, \citet{carbonetto05cm} have presented
an extension to the \citet{pasula03iu} model that solves the first
problem: estimating the number of clusters in the citation matching
domain.  Like us, they employ a Dirichlet process model to solve this
problem.  The fact that this model has now been proposed twice,
independently, is not surprising: citation matching is a well-known
problem that suffers from the need to estimate the number of clusters
in a data set, and the Dirichlet process excels at precisely this
task.)

\section{Supervised Clustering Model} \label{sec:model}

In this section, we describe our model for the supervised clustering
problem.  To facilitate discussion, we take our terminology and
notation from the reference matching task.  The canonical example of
this task is the CiteSeer/ResearchIndex database.  Specifically, we
assume that we are given a list of references appearing in the
bibliographies of scholarly publications and that we need to identify
which references correspond to the same publication.  This task is
difficult: according to CiteSeer, there are currently over 100
different books on \emph{Artificial Intelligence} by Russell and
Norvig, according to \citet{pasula03iu}.  We refer to the set $\cX$ as
the set of \emph{references} and a correct cluster of references as a
\emph{publication}.  In our problem, the observed data is a set of
references paired with partial equivalence classes over those
references (partial publications).  For instance, we might know that
$r_1,r_2,r_3 \in \cX$ belong to the same equivalence class (are the
same publication), but we might not have any information about the
equivalence class of $r_4$.  In this case, we identify $r_1,r_2,r_3$
as training data and $r_4$ as test data.

In general, we have a countable set of references $\cX$ and some
information about the structure of equivalence classes on this set and
seek to extend the observed equivalence classes to all of $\cX$.  In
complete generality, this would be impossible, due to the infinite
nature of $\cX$ and the corresponding equivalence classes.  However,
in the \emph{prediction} case, our job is simply to make predictions
about the structure of a \emph{finite} subset of $\cX$, which we have
previously denoted $X\noth$.  Thus, while our inference procedure
attempts to uncover the structure of an infinite structure,
calculations are possible because at any given time, we only deal with
a finite portion of this set.  This is not unlike the situation one
encounters in Gaussian processes, wherein a distribution is placed
over a function space, but computations are tractable because
observations are always finite.


\subsection{Generative Story} \label{sec:genstory}

The model we describe is a generative one.  Our modeling assumption is
that a reference is generated according to the cross-product of two
attributes.  The first attribute specifies which publication this
reference belongs to.  The second attribute specifies the manner in
which this reference is created, which we call the ``reference type.''
A reference type encompasses the notion that under different
circumstances, references to the same publication are realized
differently.

In the terminology of reference matching, in the context of a short
workshop paper (for instance), author first names might be abbreviated
as initials, page numbers might be left off and conferences and
journals might be referred to by abbreviations.  On the contrary, in a
reference appearing in a journal, page numbers are included, as are
full conference/journal names and author names.  In the context of
coreference resolution, one reference type might be for generating
proper names (``Bill Clinton''), one for nominal constructions (``the
President'') and one for pronouns (``he'').  Of course, the form and
number of the reference types is unknown.

The generative process for a data set proceeds as follows:

\begin{enumerate}
\item Select a distribution $G_0^p$ over publications that will be
  referred to in this data set.  $G_0^p$ should assign positive
  probability to only a finite set of all possible publications.

\item Select a distribution $G_0^t$ over reference types that will be
  used in this data set; again, $G_0^t$ should be finite.

\item For each reference $r_n$ appearing in the data set:
\begin{enumerate}
\item Select the corresponding publication $p_n \sim G_0^p$.
\item Select the corresponding reference type $t_n \sim G_0^t$.
\item Generate $r_n$ by a problem-specific distribution parameterized
  by the publication and reference type: $r_n \sim F(p_n,t_n)$.
\end{enumerate}
\end{enumerate}

The difficulty with this model is knowing how to parameterize the
selection of the distributions $G_0^p$ and $G_0^t$ in steps 1 and 2.
It turns out that a Dirichlet process is an excellent tool for solving
this problem.  The Dirichlet process (DP), which is a
\emph{distribution over distributions}, can be most easily understood
via a generalized P\`olya urn scheme, where one draws colored balls
from an urn with replacement.  The difference is that when a black
ball is drawn, one replaces it together with a ball of a new color.
In this way, the number of ``classes'' (ball colors) is unlimited, but
defines a discrete distribution (with probability one).  See Appendix
A for a brief review of the properties of the DP that are relevant to
our model.

Our model is seen as an extension of the standard na\"ive-Bayes
multiclass classification model (in the Bayesian framework), but where
we allow the number of classes to grow unboundedly.  Just as a
multiclass classification model can be seen as a finite mixture model
where the mixture components correspond to the finite classes, the
supervised clustering model can be seen as an \emph{infinite} mixture
model.  In the case of the standard multiclass setup, one treats the
class $y$ as a random variable drawn from a multinomial distribution
$\Mult(\vec \pi)$, where $\vec \pi$ is again a random variable with
prior distribution $\Dir(\vec \al)$ for the standard Dirichlet
distribution.  In our model, we essentially remove the requirement
that there is a known finite number of classes and allow this to grow
unboundedly.  In order to account for the resulting
non-identifiability of the classes, we introduce the notion of
reference types to capture the relationships between elements from the
same class.

Whenever one chooses a model for a problem, it is appropriate to
ascertain whether the chosen model is able to adequately capture the
required aspects of a data set.  In the case of our choice of the
Dirichlet process as a prior over publications, one such issue is that
of the expected number of publications per citation.  We have
performed such experiments and verified that on a variety of problems
(reference matching, identity uncertainty and coreference resolution),
the Dirichlet process is appropriate with respect to this measure (see
Section~\ref{sec:appropriateness} and Figure~\ref{fig:dp-alpha} for
discussion).

\subsection{Hierarchical Model} \label{sec:hierarchical-model}

The model we propose is structured as follows:

\begin{hierarchical2col} \label{eq:scm-hier}
\vec \pi^p & \al^p      & \Dir({\al^p}/K, \dots, {\al^p}/K)&
\vec \pi^t & \al^t      & \Dir({\al^t}/L, \dots, {\al^t}/L)\\
c_n        & \vec \pi^p & \Disc(\pi^p_1, \dots, \pi^p_K)&
d_n        & \vec \pi^t & \Disc(\pi^t_1, \dots, \pi^t_L)\\
p_k        & G^p_0      & G^p_0 &
t_k        & G^t_0      & G^t_0 \\
r_n        & c_n,d_n,\vec p, \vec t & F(p_{c_n}, t_{d_n})
\end{hierarchical2col}

\begin{figure}
\center
\psfig{figure=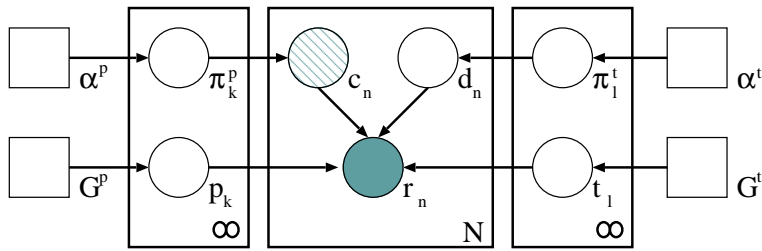,width=4in}
\caption{Graphical model for our generic supervised clustering model.}
\label{fig:gm-scm}
\end{figure}

The corresponding graphical model is depicted in
Figure~\ref{fig:gm-scm}.  In this figure, we depict the $\al$ and
$G$ parameters as being fixed (indicated by the square boxes).  The
$\al$s give rise to multinomial random variables $\pi$, which in turn
determine indicator variables $c_n$ (specifying the publication to
which $r_n$ belongs) and $d_n$ (specifying the reference type used by
reference $r_n$).  The base density $G^p$ generates publications
$p_k$ (according to a problem-specific distribution), while the base
density $G^t$ generates reference types $t_l$ (again according to a
problem-specific distribution).  Finally, the observed reference $r_n$
is generated according to publication $p_{c_n}$ and reference type
$t_{d_n}$ with problem-specific distribution $F$.  The $r_n$ random
variable (the reference itself) is shaded to indicate that it is
always observed, and the $c_n$ random variable (the indicator as to
which publication is used for reference $r_n$) is partially shaded to
indicate that it is sometimes observed (in the training data) and
sometimes not (in the test data).

As indicated by the counts on the plates for the $(\pi^p,p)$ and
$(\pi^t,t)$ variables, we take the limit as $K \fto \infty$ and $L
\fto \infty$ (where $K$ is the number of publications and $L$ is the
number of reference types).  This limit corresponds to a choice of a
Dirichlet process prior on the $p$s and $t$s \citep{neal-TR1998}.

\section{Inference Scheme} \label{sec:inference}

Inference in infinite models differs from inference in finite models,
primarily because we cannot store all possible values for infinite
plates.  However, as noted earlier, we only encounter a finite amount
of data, so at any time only a finite number of these infinite
parameters will be active---i.e., only a finite number of them will
affect the distribution of the observed data.  We will suggest and
implement inference schemes based on Markov chain Monte Carlo (MCMC)
techniques, which are the most frequently used methods for inference
in DP models
\citep{antoniak74mixtures,escobar-JASA1994,neal-TR1998,maceachern-JCGS1998,ishwaran-JASA2001,beal02infinite,xing04haplotype}.
Recently, \citet{blei05variational} have presented a variational
approach to Dirichlet process models, and \citet{minka04dpep} have
presented an inference procedure for DP models based on expectation
propagation.  Unfortunately, these methods do not work when the prior
distributions $G_0$ are not conjugate to the data distribution $F$ and
they are thus not of use to us.

The MCMC-based Bayesian solution to the supervised clustering problem
(or, indeed, any problem) is to write down the expression
corresponding to the posterior distribution of the $c_n$s for the test
data and draw samples from that posterior.  Writing data points $1$
through $N$ as the training data and points $N+1$ through $N+M$ as the
test data, we obtain the following expression for this posterior (the
actual distributions are from Eq~\eqref{eq:scm-hier}):

\begin{eqnarray*} 
&& \hspace{-7mm}
\p{\vec c_{N+1:N+M} \| \vec r_{1:N+M}, \vec c_{1:N}} \propto
  \int \ud \vec \pi^p ~
  \p{\vec \pi^p \| \al^p}
  \int \ud \vec \pi^t~
  \p{\vec \pi^t \| \al^t} \\ 
&& \hspace{5mm}
  \int \ud \vec p    ~ 
  \p{\vec p     \| G_0^p}
  \int \ud \vec t~
  \p{\vec t     \| G_0^t}
  \sum_{\vec d_{1:N+M}}
  \prod_{n=1}^{N+M} 
    \p{c_n \| \vec \pi^p}
    \p{d_n \| \vec \pi^t}
    \p{r_n \| p_{c_n}, t_{d_n}} \nonumber
\end{eqnarray*}

We now describe how we can do this sampling.  Most of the information
in this section is taken from \citet{neal-TR1998}, in which a vast
amount of additional information is provided.  The interested reader
is directed there for additional motivation and different algorithms.
The algorithms we use in this paper are either exact replicas, or
slight deviations from Algorithms 2 and 8 of Neal's.

\subsection{Updates for Conjugate Priors}

\begin{figure}[t]
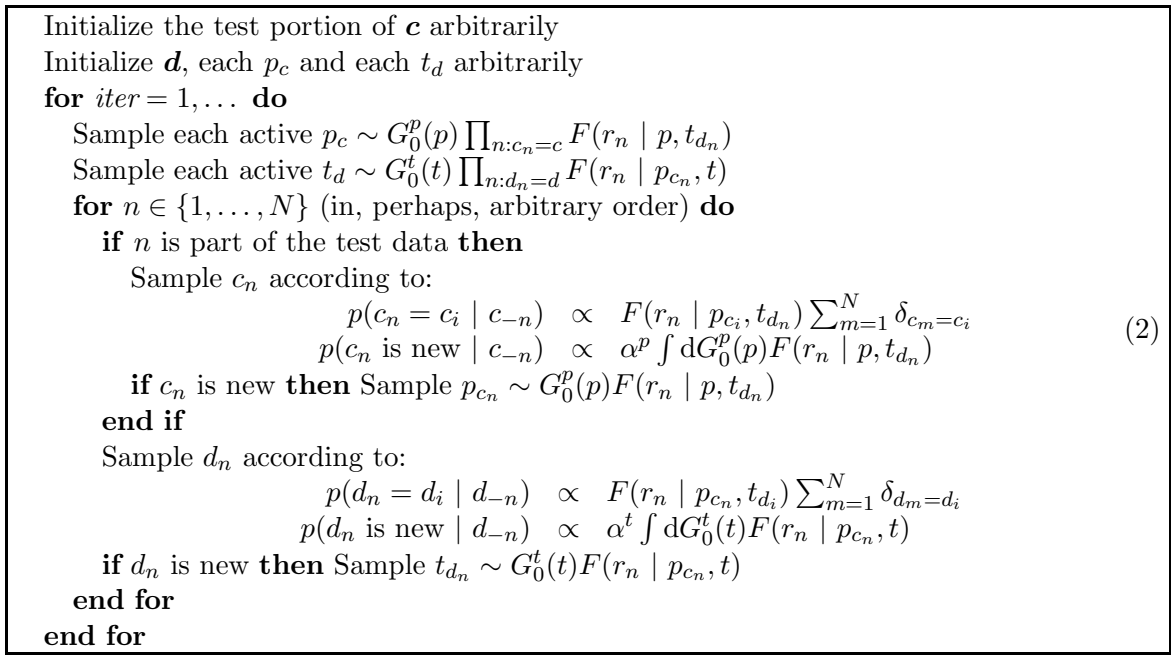

\framebox{\begin{minipage}[t]{\textwidth}
\begin{algorithmic}
\STATE Initialize the test portion of $\vec c$ arbitrarily
\STATE Initialize $\vec d$, each $p_c$ and each $t_d$ arbitrarily
\FOR{$\textit{iter} = 1, \dots$}
\STATE Sample each active $p_c \by G_0^p(p) \prod_{n : c_n = c} F(r_n \| p, t_{d_n})$
\STATE Sample each active $t_d \by G_0^t(t) \prod_{n : d_n = d} F(r_n \| p_{c_n}, t)$
\FOR{$n \in \{ 1, \dots, N \}$ (in, perhaps, arbitrary order)}
\IF{$n$ is part of the test data}
\STATE {Sample $c_n$ according to:
\vspace{-4mm}\begin{equation} \label{eq:sample-c}
\begin{array}{rcl}
p(c_n = c_i \| c_{-n})  &\propto&
  F(r_n \| p_{c_i}, t_{d_n}) \sum_{m=1}^N \de_{c_m = c_i}
                  \\
p(c_n \textrm{ is new} \| c_{-n}) &\propto&
  \al^p \int \ud G_0^p(p) F(r_n \| p, t_{d_n})
\end{array}
\end{equation}}
\vspace{-6mm}
\STATE {\bf if} $c_n$ is new {\bf then} Sample $p_{c_n} \by G_0^p(p) F(r_n \| p, t_{d_n})$
\ENDIF
\STATE {Sample $d_n$ according to:
\vspace{-4mm}\begin{equation*} 
\begin{array}{rcl}
p(d_n = d_i \| d_{-n})  &\propto&
  F(r_n \| p_{c_n}, t_{d_i}) \sum_{m=1}^N \de_{d_m = d_i}
                  \\
p(d_n \textrm{ is new} \| d_{-n}) &\propto&
  \al^t 
             \int \ud G_0^t(t) F(r_n \| p_{c_n}, t)
\end{array}
\end{equation*}}
\vspace{-6mm}
\STATE {\bf if} $d_n$ is new {\bf then} Sample $t_{d_n} \by G_0^t(t) F(r_n \| p_{c_n}, t)$
\ENDFOR
\ENDFOR
\end{algorithmic}
\end{minipage}}
\caption{The inference algorithm for the supervised clustering model
  with conjugate priors.}
\label{fig:full-bayes}
\end{figure}

The simplest case arises when a conjugate prior is used.  In the
terminology of the Dirichlet process, this means that the data
sampling distribution $F$ is conjugate to the base density $G_0$ of
the Dirichlet process.  To perform inference with conjugate priors, we
need to be able to compute the marginal distribution of a single
observation and need to be able to draw samples from the posterior of
the base distributions.  In each iteration of sampling, we first
resample each active publication $p_c$ and reference type $t_d$
according to their posterior densities (in the case of conjugate
priors, this is possible).  Then, for each test reference, we resample
its publication and for all references, we resample the corresponding
reference type.  The algorithm is shown in
Figure~\ref{fig:full-bayes}.  We actually have two options when
sampling the $c_n$s, depending on whether publications are allowed to
be shared across the training and testing data.  If a training
reference may refer to the same publication as a testing reference (as
is natural in the context of reference matching), then the sum in
Eq~\eqref{eq:sample-c} is over all data; on the other hand, if they
are not allowed to co-refer (as is natural in, for example,
single-document coreference resolution), then the sum is only over the
test data.

\subsection{Updates for Non-Conjugate Priors} \label{sec:update-nonconj}

The case of non-conjugate priors is a bit more complex, since in this
case, in general, one is not able to analytically compute the data
marginals, nor is one able to directly sample from the relevant
posterior distributions.  A na\"ive solution would be to set up
separate Markov chains to draw samples from the appropriate
distributions so that we \emph{could} calculate these.  Unfortunately,
since these values need to be computed for each loop of the ``outer''
Markov chain, such an approach is impractical.  The
alternative---given as Algorithm 8 by \citet{neal-TR1998}---is
essentially to sample just a few of these needed values in a way that
does not affect the detailed balance condition that guarantees that
the \emph{outer} Markov chain converges to the correct stationary
distribution.

The overall structure of the sampling algorithm remains identical in
the case of non-conjugate priors; however, the sampling for the
indicator variables $c_n$ and $d_n$ changes slightly, and so does the
sampling of the $p$ and $t$ variables.  For instance, in the conjugate
case, $d_n$ is sampled according to the marginal distribution $\int
\ud G_0^t(t) F(r_n \| p_{c_n}, t)$, which is analytically unavailable
when $G_0^t$ is not conjugate to $F$ (with respect to the second
variable).  In the case of non-conjugacy, we approximate this integral
by drawing $\tilde M$ samples independently from $G_0^t$.  In general,
as $\tilde M \fto \infty$, this is exactly like computing the integral
with an independence sampler; however, for $\tilde M$ finite, we still
get convergence of the overall Markov chain.  $\tilde M$ is set by the
experimenter by choosing the number of samples $M$ that is drawn and
then setting $\tilde M$ to be $M$ whenever the old value of $d_n$ was
not unique, and to $M+1$ whenever it was unique.  If the chosen value
corresponds to one of the newly sampled $t$s, then we set $t_d$ to be
that sampled value.  The corresponding sampling for the $c$ variables
is identical.  This is the technique suggested by \citet{neal-TR1998}
in his Algorithm 8.  In all experiments, we use $M=8$.

The second complication is when we cannot sample from the data
posteriors, which means that resampling $p$ and $t$ is difficult.
This is partially assuaged by the fact that in sampling for $c_n$ and
$d_n$ we are given an explicit new value of $p$ or $t$ to use.
However, at the beginning of each iteration of the chain, we must
resample $p$ according to its posterior distribution (and similarly
for $t$).  The most general approach to solving this problem---and
the approach we employ here---is to run a short independence sampler
for $p$ by drawing a set of values $p$ from $G_0^p$ and then choosing
one of those according to its posterior.  However, depending on the
actual distributions chosen, there might be more appropriate methods
for doing this sampling that still leaves the overall chain invariant.

\subsection{Resampling the Dirichlet Process Precision} \label{sec:sample-al}

We often wish to leave the values of $\al^p$ and $\al^t$ (the
scaling/precision hyperparameters for the two Dirichlet processes) as
random variables, and estimate them according to the data
distribution.  \citet{west92hyperparam} gives a method for drawing
samples for the precision parameter given the number of references
$N$and the number of publications $K$ (or, for $\al^t$, the number of
reference types); in his analysis, it is natural to place a gamma
prior on $\al$.  In most cases, his analysis can be applied directly;
however, in the case of coreference resolution, the problem is a bit
more complicated because we have \emph{multiple} observations pairs
$(N,K)$ for each ``training document.''  In Appendix B, we briefly
extend this analysis to the case where there are multiple
observations.

\section{Model Parameterization} \label{sec:param}

One of the simplest model parameterizations occurs when the data
points $r_n$ are vectors in the Euclidean space $\R^F$ for some
dimensionality $F$, and when each dimension is a measure of distance
(i.e., $|r_{nf} - r_{mf}|$ is small whenever $r_n$ and $r_m$ are
similar along dimension $f$).  In this case, it may be a reasonable
assumption that the $r_n$s are distributed normally around some
unknown mean vector, and with some unknown covariance.  While the
assumption of normalcy is probably not accurate, it turns out that it
fares rather well experimentally (see Section~\ref{sec:results}).
Moreover, as discussed at the end of this paper, it is possible to
substitute in other models for $F$ as deemed appropriate by a specific
problem.

If we believe the $r_n$s are distributed normally (i.e., $F$ is a
Normal distribution), it is natural to treat the $p_k$ variables as
means and the $t_l$ variables as precisions (inverse
variance-covariances matrices).  For efficiency's sake, we 
further assume that $t_l$ is \emph{diagonal}, so that all covariance
terms are zero.  In this model, one can think of a precision $t_{lf}$
as the ``weight'' along dimension $f$, so that high weights mean that
this dimension is important and low weights mean that this dimension
is not relevant.

By making $F$ an isotropic Normal distribution, the natural conjugate
priors are to make $G_0^p$ another Normal distribution and to make
$G_0^t$ a product of inverse-gamma distributions (one inverse-gamma
distribution per dimension $f$).\footnote{If we had not assumed that
$t$ was diagonal, then the natural choice for $G_0^t$ would be an
inverse-Wishart distribution.}  As we typically center and spherize
the training data, it is natural to parameterize $G_0^p$ with a mean
of $0$ and a covariance matrix of $\si \mat I$ for some $\si \approx
1$.  Similarly, we may parameterize $G_0^t$ with identical scale and
shape parameters all approximately $1$.  (Note that we could also
\emph{learn} these hyperparameters during inference by including them
in the sampling, though we do not explore this option.)

Experiments with the model just described have demonstrated that while
it is adept at finding points in the same cluster, it is not as able
to separate out points in different clusters (it has low precision, in
the precision/recall sense).  This occurs because the Gaussian
precisions are learned solely for the purpose of accounting for the
distribution of classes by themselves, but with no regard to the
relation between classes.  We explore two modeling extensions to
attempt to alleviate this problem and give the model a better ability
to separate classes; in the first, we maintain conjugacy (and hence
efficiency in implementation), but in the second we give up conjugacy
for a more appropriate model.

\subsection{Separation by Modifying $G_0^p$} \label{sec:modg0}

Our first method for adding separation power between to the model is
to condition the parameters of $G_0^p$ on $\vec p$ and $\vec c$: in
other words, the shape and scale parameters of the prior on the
precisions is affected by the relative positions of the means of the
data.  In the original model, we assumed that $t_f \by \Gam(1,1)$ is a
gamma random variable with mean $1$ and variance $1$.  Here, we wish
to change this distribution so that the mean is large enough to keep
the data separated along this dimension, and the variance is small
whenever many points tell us that this dimension is important.  To
accomplish this we use instead a $\Gam(a,b)$ prior, where $ab$ is half
the mean variance along dimension $f$ and $ab^2$ is the variance of
the variance along dimension $f$.  The values for $a$ and $b$ must be
resampled at each iteration of the algorithm.

\subsection{Separation by Conditioning} \label{sec:cond}

Our second approach to adding more separation power to the model is to
condition the choice of the precisions (reference types) $t$ on the
means (publications) $p$.  In terms of our generative story, this
means that first we choose a publication then, based on the
publication, choose a reference type.  Since we wish to ascribe no
meaning to the actual location of the means $p_k$, we compute
this probability based only on their relative distances (along each
dimension), and also under a na\"ive Bayes assumption:

\begin{eqnarray}
\p{\vec t \| \vec p, \vec c, \vec d \hide{, G_0^p, G_0^t}}
&\raisebox{0pt}[0pt][0pt]{$\approx$\raisebox{4mm}{\hspace{-3mm}\scriptsize[1]}}&
\prod_{i=1}^{|\vec d|} 
  \p{t_i \| \vec p, \vec c, \vec d\hide{, G_0^p, G_0^t}}
\nonumber\\
&\raisebox{0pt}[0pt][0pt]{$=$\raisebox{4mm}{\hspace{-3mm}\scriptsize[2]}}&
\prod_{i=1}^{|\vec d|} 
  \frac {\p{t_i \| \vec c, \vec d\hide{, G_0^p, G_0^t}}
         \p{\vec p \| t_i, \vec c, \vec d\hide{, G_0^p, G_0^t}}}
        {\p{\vec p \| \vec c, \vec d\hide{, G_0^p, G_0^t}}}
\nonumber\\
&\raisebox{0pt}[0pt][0pt]{$\approx$\raisebox{4mm}{\hspace{-3mm}\scriptsize[3]}}&
\prod_{i=1}^{|\vec d|} 
  \frac {G_0^t(t_i)} {\prod_{j=1}^{|\vec c|} G_0^p(p_j)}
  \prod_{j=1}^{|\vec c|}
    \p{p_j \| t_i, \vec p_{1:j-1}, \vec c, \vec d\hide{, G_0^p, G_0^t}}
\nonumber\\
&\raisebox{0pt}[0pt][0pt]{$=$\raisebox{4mm}{\hspace{-3mm}\scriptsize[4]}}&
\prod_{i=1}^{|\vec d|} 
  G_0^t(t_i)
  \prod_{j=1}^{|\vec c|}
    \frac {
      \p{p_j \| t_i, \vec c, \vec d\hide{, G_0^p, G_0^t}}
      \p{\vec p_{1:j-1} \| t_i, p_j, \vec c, \vec d\hide{, G_0^p, G_0^t}}
      }
      { \p{\vec p_{1:j-1} \| t_i, \vec c, \vec d\hide{, G_0^p, G_0^t} }
        G_0^p(p_j) 
      }
\nonumber\\
&\raisebox{0pt}[0pt][0pt]{$\approx$\raisebox{4mm}{\hspace{-3mm}\scriptsize[5]}}&
\prod_{i=1}^{|\vec d|} 
  G_0^t(t_i)
  \prod_{j=1}^{|\vec c|}
    \frac {
      \p{p_j \| t_i, \vec c, \vec d\hide{, G_0^p, G_0^t}}
      \prod_{k=1}^{j-1}
      \p{p_k \| t_i, p_j, \vec c, \vec d\hide{, G_0^p, G_0^t}}
      }
      { G_0^p(p_j)
        \prod_{k=1}^{j-1} \p{p_k \| t_i, \vec c, \vec d\hide{, G_0^p, G_0^t}}
         }
\nonumber\\
&\raisebox{0pt}[0pt][0pt]{$=$\raisebox{4mm}{\hspace{-3mm}\scriptsize[6]}}&
   \prod_{i=1}^{|\vec d|}
     G_0^t(t_i)
     \prod_{j=1}^{|\vec c|}
       G_0^p(p_i)^{2(j-1) - |\vec c|}
       \prod_{k=1}^{j-1} \p{p_j \| p_k, t_i}
  \label{eq:condition}
\end{eqnarray}

In the first step of this derivation, we make a factorial assumption
on the $\vec t$ vector.  The second step simply applies Bayes' rule.
The third step replaces the generic $\p{\cdot}$ symbol for the $t_i$
variables with the true distribution $G_0^t$, makes a similar
factorial assumption on the $\vec p$ vector and replaces the
corresponding $\p{\cdot}$ with $G_0^p$.  The fourth step applies
Bayes' rule to the last term and moves the denominator from the first
product into the second.  The fifth step applies the same factorial
assumption on $\vec p_{1:j-1}$ as before.  The last step replaces the
generic $\p{\cdot}$ symbol with $G_0^p$ and performs some minor
algebraic manipulation.


This final expression in Eq~\eqref{eq:condition} depends only on the
prior values for the sampled $\vec t$s and $\vec p$s, coupled with the
probability of mean $p_j$ given $p_k$ under precision $t_i$.
Unfortunately, under the assumptions made, the probability of a vector
$\vec p$ is no longer independent of the ordering of the values of
$\vec p$.  In all our experiments, we order the $\vec p$ according to
the sizes of the classes: if $\textit{count}(c_1) >
\textit{count}(c_2)$.  We parameterize the distribution on the means
$\p{p_j \| p_k, t_i}$ by treating the \emph{distance} between $p_j$
and $p_k$, measured by $t_i$ as a random variable with an exponential
distribution: $\p{p_j \| p_k, t_i} = \la \exp [-\la \norm{p_j -
p_k}^2_{t_i}]$.  We set $\la = 1$, but, again, it could be learned
concurrently by sampling.

Clearly, this prior distribution for $t$ is no longer conjugate to the
data sampling distribution $F$.  Moreover, the $\vec p$s and $\vec t$s
are no longer separated by the indicator variables, which makes the
entire sampling story more complex.  Indeed, the marginal distribution
now depends on the types and, similarly, the types depend on the
mentions.  We thus use the non-conjugate updates described in
Section~\ref{sec:update-nonconj}.  The simplest approach to performing
inference with the non-conjugate priors would be, for each of the
$\tilde M$ samples for $\vec p$, to draw from $G_0^p$ and weight the
sampled $\tilde p$s proportional to its unnormalized posterior
probability, given by Eq~\eqref{eq:condition}.  Similarly, a proposed
sample $\tilde t$ would be weighted according to its (unnormalized)
posterior probability according to Eq~\eqref{eq:condition}.

\section{Performance Metrics} \label{sec:metrics}

Quite a few performance metrics have been proposed in the literature
for comparing two clusterings of a given data set.  Since these are,
in general, less well known than the metrics used for classification
(accuracy, ROC, etc.), we review them here, and attempt to point
out the strengths and weaknesses of each metric.  Of course, the
evaluation criteria one uses should reflect one's own personal views
of what is important, but the metrics used here can be seen as
surrogate measurements when such prior knowledge is unavailable.  All
of these metrics assume that we have a gold standard (correct)
clustering $G$ and a hypothesis clustering $H$ and that the total
number of data points is $N$.

\subsection{Rand Index}

The rand index \citep{rand71index} is computed by viewing the
clustering problem as a binary classification problem.  Letting
$N_{11}$ denote the number of pairs that are in the same cluster in
both $G$ and in $H$, and letting $N_{00}$ denote the number of pairs
that are in different clusters in both $G$ and $H$, the rand index has
value $\textnormal{\bsf{RI}}(G,H) = 2 [N_{11} + N_{00}]/[N(N-1)]$.
Thus, the rand index computes the number of correct binary decisions
($N_{11}+N_{00}$) made by the system and normalizes by the total
number of decisions made.  The value of the rand index lies between
$0$ and $1$, with $1$ representing a perfect clustering.

The rand index is the most frequently reported metric in the
clustering literature, though we believe that its value is often
misleading.  As we show in our results (Section~\ref{sec:results}), a
very simple baseline system that places each element in its own
cluster tends to achieve a very high rand index.  This occurs due to
the structure of the clusters in most real world data sets.  In such
data sets, the number of negative pairs (pairs that, in the gold
standard, fall into different clusters) vastly outnumber the number of
positive pairs; thus the rand index becomes dominated by the $N_{00}$
factor, and the $N_{11}$ factor tends to have very little impact on
the final value.  Moreover, the influence of large clusters on the
rand index quadratically outnumbers the influence of small clusters on
this value, so system performance on small clusters (which are
typically the most difficult) becomes insignificant.

In this paper, we report the rand index for comparative purposes with
earlier work, but strongly encourage readers not to take these numbers
too seriously.  We recommend other researchers in the supervised
clustering field to report on other metrics of system performance than
the rand index.

\subsection{Precision, Recall, F-score}

The second set of metrics we report are the precision/recall/F-score
of the clustering.  Extending the notation used for the rand index, we
write $N_{10}$ for the number of pairs that are in the same
cluster in $G$, but in different clusters in $H$.  Similarly, we 
write $N_{01}$ for the number of pairs that are in different clusters
in $G$ but the same cluster in $H$.  Precision is
$\textnormal{\bsf{P}}(G,H) = N_{11} / [N_{11} + N_{01}]$, recall is
$\textnormal{\bsf{R}}(G,H) = N_{11} / [N_{11} + N_{10}]$ and F-score
is $\textnormal{\bsf{F}}(G,H) = (\textnormal{\bsf{P}}(G,H)^{-1} +
\textnormal{\bsf{R}}(G,H)^{-1})^{-1}$.  Again, each of these values
falls between $0$ and $1$ with $1$ being optimal.  While precision,
recall and F-score are still computed based on binary 
decisions,
they do not suffer as strongly from the weaknesses of the rand index.
However, they still place quadratically as much importance on large
clusters.

\subsection{Cluster Edit Distance and Normalized Edit Score}

\citet{pantel-thesis} proposes a metric called the \emph{cluster edit
distance}, which computes the number of ``create,'' ``move,'' and
``merge'' operations required to transform the hypothesis clustering
into the gold standard.  Since no ``split'' operation is allowed, the
cluster edit distance can be computed easily and efficiently.
However, the lack of a split operation (which is absent precisely so
that the computation of the metric is efficient) means that the
cluster edit distance favors algorithms that tend to make too many
clusters, rather than too few clusters.  This is because if an
algorithm splits an $m$ element cluster in half, it requires only one
merge operation to fix this; however, if, instead, two $m/2$-sized
clusters are mistakenly merged by an algorithm, $m/2$ operations are
required to fix this error.  The cluster edit distance has a minimum
at $0$ for the perfect clustering and a maximum of $N$.  Also note
that the cluster edit distance is not symmetric: in general, it does
not hold that $\textnormal{\bsf{CED}}(G,H) = \textnormal{\bsf{CED}}(H,G)$ (again, precisely
because splits are disallowed).

We propose a variant of the cluster edit distance that we call the
\emph{normalized edit score}.  This value is computed as
$\textnormal{\bsf{NES}}(G,H) = 1 - [\textnormal{\bsf{CED}}(G,H) +
\textnormal{\bsf{CED}}(H,G)] / [2N]$ and is clearly symmetric and no
longer favors fine clusterings over coarse clusterings.  Additionally,
it takes values from $0$ to $1$, with $1$ being a perfect clustering.
While the normalized edit score no longer can be interpreted in terms
of the number of operations required to transform the hypothesis
clustering into the correct clustering, we believe that these
additional properties are sufficiently important to make it
preferable to the cluster edit distance metric.

\subsection{Variation of Information}

The final metric we report in this paper is the variation of
information (\bsf{VI}), introduced by \citet{meila03clusterings}.  The
\bsf{VI} metric essentially looks at how much entropy there is about
$G$ knowing $H$, and how much entropy there is about $H$ knowing $G$.
It is computed as $\textnormal{\bsf{VI}}(G,H) = H(G) + H(H) -
2I(G,H)$.  Here, $H(\cdot)$ is the entropy of a clustering, computed
by looking at the probability that any given point is in any
particular cluster.  $I(G,H)$ is the mutual information between $G$
and $H$, computed by looking at the probability that two points are in
the same cluster, according to $G$ and $H$.  It has a minimum at
$0$, only when the two clusterings match, and is bounded above by
$\log N$.  It has several other desirable properties, including the
fact that it is a metric.  Though frowned upon by
\citet{meila03clusterings}, we also report the \emph{normalized
variation of information}, computed simply as:
$\textnormal{\bsf{NVI}}(G,H) = 1 - \textnormal{\bsf{VI}}(G,H) / \log
N$.  This value is again bounded between $0$ and $1$, where $1$
represents a correct clustering.

\section{Experimental Results} \label{sec:results}

In this section, we present experimental results on both artificial
and real-world data sets, comparing our model against other supervised
clustering algorithms as well as other standard clustering
algorithms.  We first discuss the baselines and systems we
compare against, and then describe the data sets we use for
comparison.  Some data sets support additional, problem-specific
baselines against which we also compare.

\subsection{Systems Compared} \label{sec:results:systems}

The first baseline we compare against, \system{Coarse}, simply places
all elements in the same, single cluster.  The second baseline,
\system{Fine}, places each element in its own cluster.  These are
straw-man baselines that are used only to provide a better sense of
the performance metrics.

The next systems we compare against are pure clustering systems that
do not perform any learning.  In particular, we compare against
\system{K-Means}, where the number of clusters, $k$, is chosen
according to an oracle (this is thus an \emph{upper bound} on how well
the k-means algorithm can perform in real life).  We additionally
compare against a version of our model that does not use any of the
training data.  To do so, we initialize $\al^p = 1$ and use a single
reference type, the identity matrix.  This system is denoted
\system{CDP} (for ``{\bf C}lustering with the {\bf D}richlet {\bf
P}rocess'') in subsequent sections.

The final class of systems against which we compare are true learning
systems.  The first is based on the standard technique of building a
binary classifier and applying a clustering method to it.  We use an
SVM as the classifier, with an RBF kernel.  The kernel parameter $\ga$
and the regularization parameter $C$ are tuned using golden section
search under $10$-fold cross validation.  After the SVM has been
optimized, we use an agglomerative clustering algorithm to create
clusters according to either minimum, maximum or average link, with a
threshold to stop merging.  The link type (min, max or avg) and the
threshold is tuned through another series of $10$-fold cross
validation on the training data.  This is essentially the method
advocated by \citet{cohen02match}, with the slight complication that
we consider all link types, while they use average link exclusively.
This system is denoted \system{Binary} in subsequent sections.

The second learning system is the model of distance metric learning
presented by \citet{xing03metric}.  This model learns a distance
metric in the form of a positive semi-definite matrix $\mat A$ and
computes the distance between vectors $\vec x$ and $\vec y$ as $[(\vec
x - \vec y)\T \mat A(\vec x - \vec y)]^{1/2}$.  The matrix is learned
so as to minimize the distances between elements in the same cluster
(in the training data) and maximize the distance between elements in
different clusters.  Once this distance metric is learned,
\citet{xing03metric} apply standard k-means clustering to the test
data.  There is a weighting term $C$ that controls the trade-off
between keeping similar points close and dissimilar points separate;
we found that the performance of the resulting system was highly
sensitive to this parameter.  In the results we present, we ran four
configurations, one with $C=0$, one with $C=1$, one with $C=|s|/|d|$
(where $s$ is the set of similar points and $d$ is the set of
dissimilar points), and one with $C=(|s|/|d|)^2$.  We evaluated all
four and chose the one that performed best on the test data according
to F-score (using an ``oracle'').  In all cases, either $C=0$ or
$C=(|s|/|d|)^2$ performed best.  We denote this model \system{Xing-K}
in the following.

Lastly, we present results produced by the system described in this
paper.  We report scores on several variants of our ``{\bf S}upervised
{\bf C}lustering with the {\bf D}richlet {\bf P}rocess'' model:
\system{SCDP-1} is the result of the system run using the conjugate
inference methods; \system{SCDP-2} is the model presented in
Section~\ref{sec:modg0} that is aimed at achieving better class
separation by modifying $G_0^p$; finally, \system{SCDP-3} is the model
presented in Section~\ref{sec:cond} that separates classes through
conditioning.  For all problems, we will report the number of
iterations of the sampling algorithm run, and the time taken for
sampling.  In all cases, we ran the algorithms for what we \emph{a
priori} assumed would be ``long enough'' and did not employ any
technique to determine if we could stop early.

\subsection{Data Sets}

We evaluate these models on four data sets, the first of which is
semi-artificial, and the last three of which are real-world data sets
from three different domains.  The four data sets we experiment on
are: the \citet{usps-digits}, a collection of annotated data for
identity uncertainty from \citet{doan04ontology}, proper noun
coreference data from NIST and reference matching data from
\citet{mccallum00matching}.  In the digits data set, the data points
live in a high-dimensional Euclidean space and thus one can directly
apply all of the models discussed above.  The last three data sets all
involve textual data for which an obvious embedding in Euclidean space
is not available.  There are three obvious approaches to dealing with
such data.  The first is to use a Euclidean embedding technique, such
as multidimensional scaling, kernel PCA or LLE, thus giving us data in
Euclidean space to deal with.  The second is to modify the Gaussian
assumption in our model to a more appropriate, problem-specific
distribution.  The third, which is the alternative we explore here, is
to notice that in all the computations required in our model, in
k-means clustering, and in the distance metric learning algorithm
\citep{xing03metric}, one never needs to compute locations but only
relative distances.\footnote{For instance, one of the common
calculations is to compute the distances between the means of two
subsets of the data, $\{a_i\}_{i=1}^I$ and $\{b_j\}_{j=1}^J$.  This
can be computed as:

\vspace{-4mm}
\begin{equation*}
\norm{\frac 1 I \sum_{i=1}^I a_i -
      \frac 1 J \sum_{j=1}^J b_j }^2 =
  \frac 1 {IJ}  \sum_{i=1}^I \sum_{j=1}^J    \norm{a_i - b_j}^2 
- \frac 1 {I^2} \sum_{i=1}^I \sum_{i'=i+1}^I \norm{a_i - a_{i'}}^2 
- \frac 1 {J^2} \sum_{j=1}^J \sum_{j'=j+1}^J \norm{b_j - b_{j'}}^2
\end{equation*}
\vspace{-2mm}

\noindent
The other relevant computations can be done similarly, and the
generalization to multidimensional inputs is straightforward.}  We
thus structure all of our feature functions to take the form of some
sort of distance metric and then use all algorithms with the implicit
embedding technique.  The choice of representation is an important one
and a better representation is likely to lead to better performance,
especially in the case where the features employed are not amenable to
our factorial assumption.  Nevertheless, results with this simple
model are quite strong, comparative to the other baseline models, and
little effort was required to ``make the features work'' in this task.
The only slight complication is that the distances need to be defined
so that large distance is correlated with different class, rather than
the other way around---this is a problem not faced in conditional or
discriminative models such as those of \citet{mccallum04coref} and
\citet{finley05sc}.

\begin{figure}
\center
\begin{tabular}{lr}
\psfig{figure=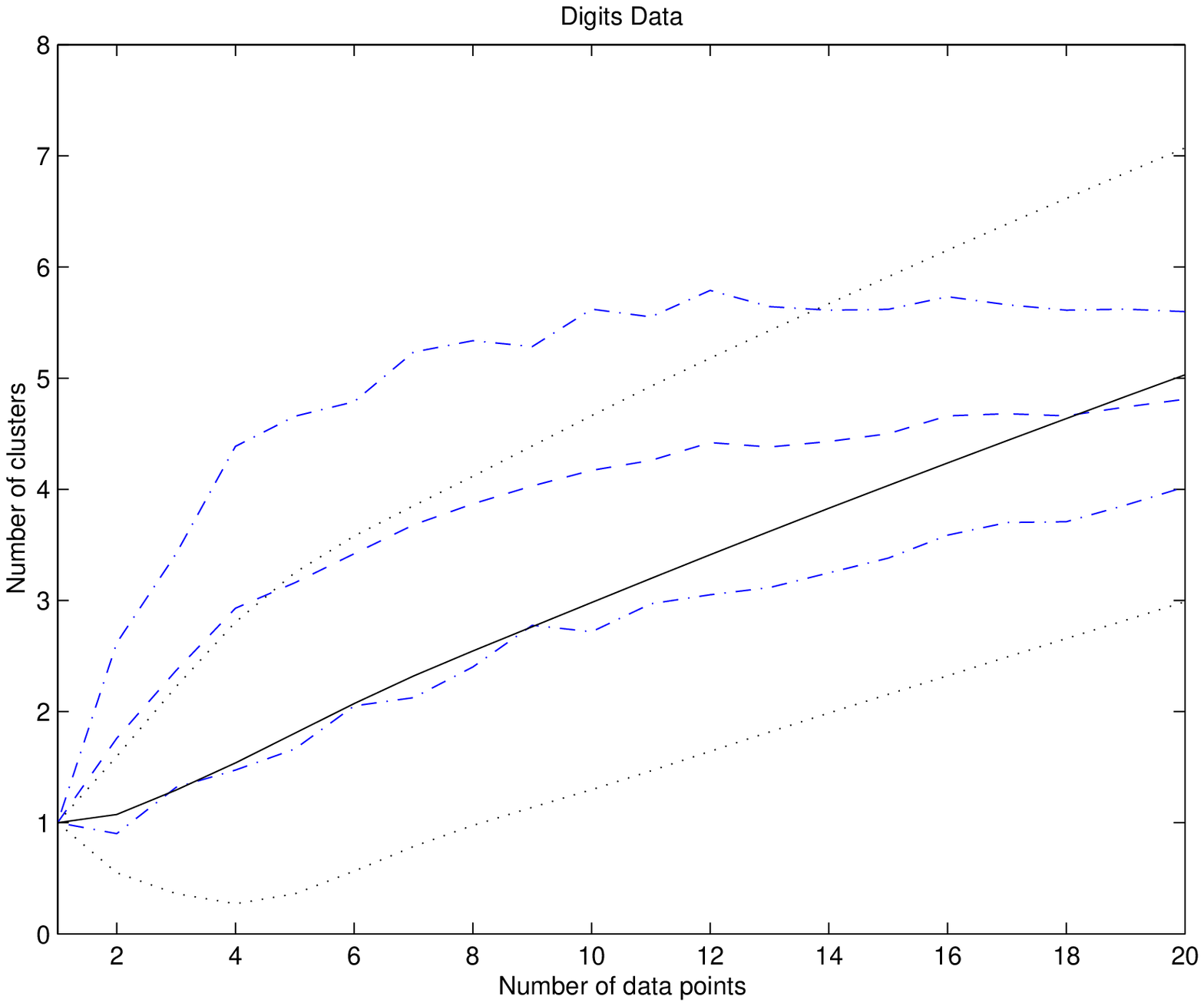,width=0.4\textwidth,height=3.9cm} &
\psfig{figure=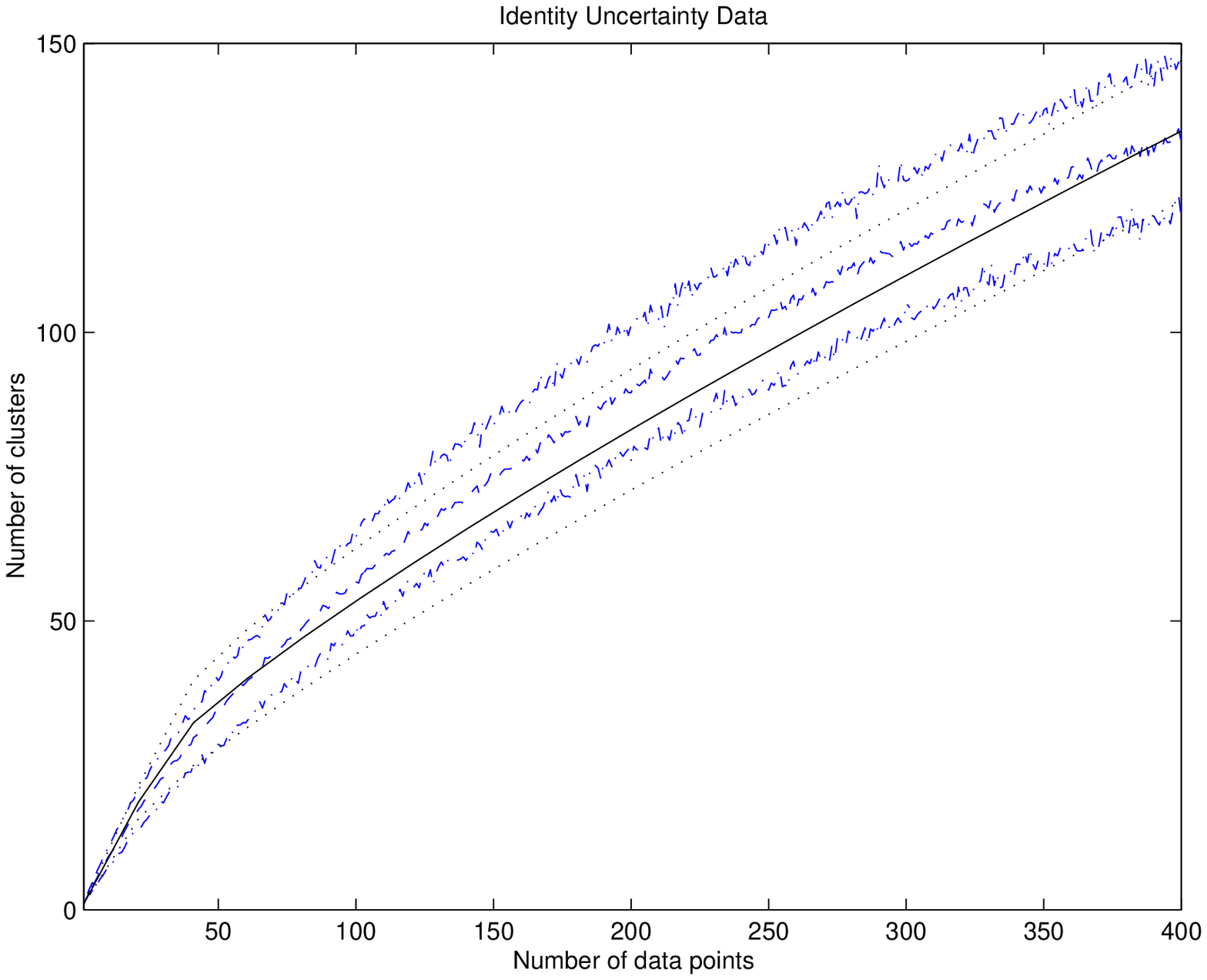,width=0.4\textwidth,height=3.9cm} \\
\psfig{figure=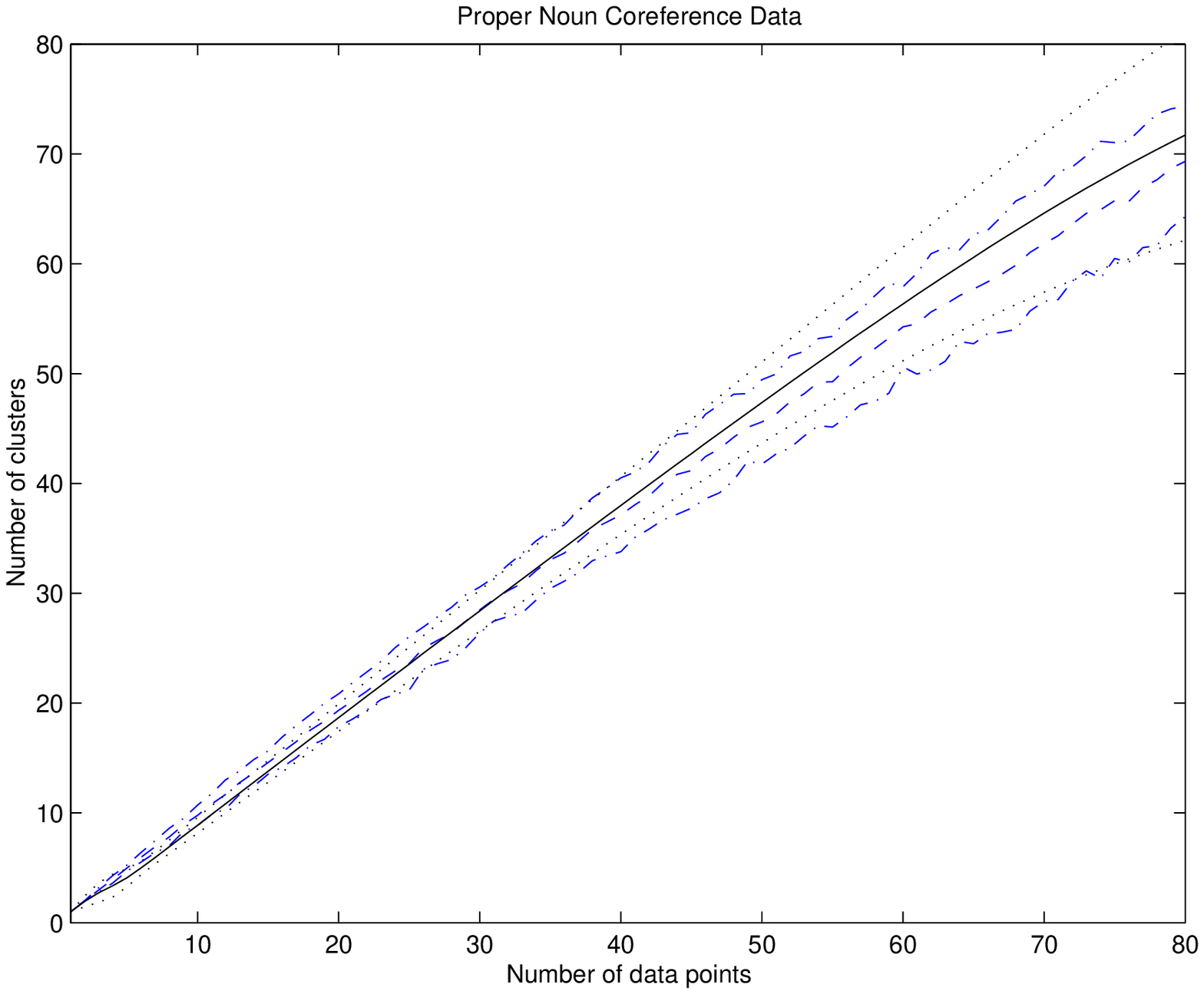,width=0.4\textwidth,height=3.9cm} &
\psfig{figure=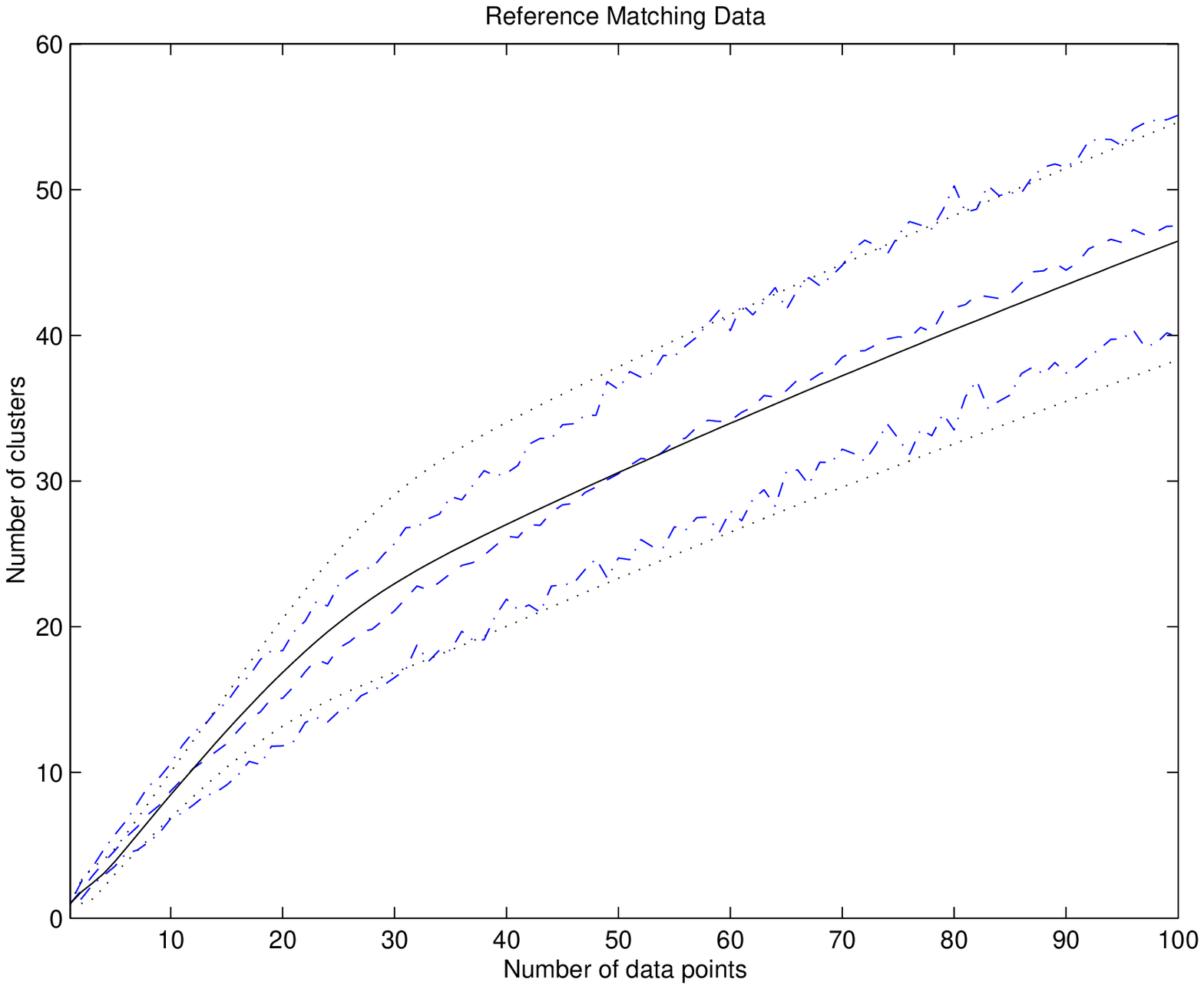,width=0.4\textwidth,height=3.9cm} 
\end{tabular}
\caption{Number of data points by expected number of clusters for the
  four data sets.  The solid black line is the expected number
  according to the Dirichlet process (dotted black lines are two
  standard deviations); the dashed blue line is the empirical expected
  number (dash-dotted blue lines are two standard deviations).}
\label{fig:dp-alpha}
\end{figure}

\subsection{Appropriateness of DP prior} \label{sec:appropriateness}

Before presenting results on these four data sets, we evaluated
whether the assumption that the underlying data distribution comes
from a Dirichlet process is reasonable.  To do so, we estimated the
$\al$ parameter for each data set as described in
Section~\ref{sec:sample-al} and computed for each data set size $N$
the expected number of classes $K$ according to the DP (as well as its
standard deviation).  For each $N$, we also computed---through
resampling---the \emph{empirical} expected value of $K$ according to
the data set and its standard deviation.  We have plotted these curves
for each data set in Figure~\ref{fig:dp-alpha}.  As we can see form
this figure, the DP is an excellent match for most of the data sets,
except for the digits data, where the match is rather poor (though the
expectations always fall within two standard deviations).  Better fits
could be obtained using a more complex prior, such as the
two-parameter Poisson-Dirichlet process, but we believe that for these
tasks, the standard DP is sufficient.

\subsubsection{Digits data}

Our first data set is adapted from the \citet{usps-digits}, originally
a test set of multiclass classification in the vision domain.  In
order to treat it as a supervised clustering problem, we randomly
selected five of the ten digits as the ``training data'' and use the
remaining five as ``test data.''  The digits used for training are
$\{1, 3, 5, 8, 9\}$ and those used for testing are $\{0, 2, 4, 6,
7\}$.  The idea is that upon seeing only the digits $\{1,3,5,8,9\}$, a
supervised clustering model should have learned enough about the
structure of digits to be able to separate the digits $\{0,2,4,6,7\}$,
even though it has seen none of them (of course, it will not be able
to label them).

In order to more closely mimic the fact that in real world data the
clusters are rarely equally sized, we artificially ``imbalanced'' both
the training and test data, so that there were between $30$ and $300$
examples of each of the digits.  The digits are $8\times 8$ blocks of
pixel intensities, which in all cases are centered and scaled to have
unit variance along each dimension.  We run ten chains of ten thousand
iterations each of the inference algorithm from
Figure~\ref{fig:full-bayes}.  Each chain of model 1 required about
$40$ minutes to complete; model 2's chains required approximately one
hour and the chains from model 3 required $5$ hours to complete.

\begin{table}
\center\small
\begin{tabular}{|l||c||c|c|c||c|c||c|c|}
\hline
\makebox[25mm][l]{\bf System} & \makebox[1cm][c]{\bsf{RI}} & \makebox[1cm][c]{\bsf{P}} & \makebox[1cm][c]{\bsf{R}} & \makebox[1cm][c]{\bsf{F}} & \makebox[1cm][c]{\bsf{CED}} & \makebox[1cm][c]{\bsf{NES}} & \makebox[1cm][c]{\bsf{VI}} & \makebox[1cm][c]{\bsf{NVI}} \\
\hline
\system{Coarse}        & .229 & .229 & 1.00 & .372 & .725 & .275 & 1.525& .765 \\
\system{Fine}          & .771 & 1.00 & .000 & .000 & 1.00 & .008 & 4.975& .235 \\
\hline
\system{K-Means}       & .760 & .481 & .656 & .555 & .350 & .412 & 1.446& .778 \\
\system{CDP}        & .886 & .970 & .016 & .031 & 1.00 & .000 & 4.237& .241 \\
\hline
\system{Binary}        & \bf .921 & .730 & .497 & .592 & .455 & .372 & 1.193& .805 \\
\system{Xing-K}        & .821 & .610 & .605 & .608 & .245 & .478 & 1.165& .821 \\
\hline
\system{SCDP-1}     & .848 & .668 & .664 & .666 & .239 & .483 & 1.176& .819 \\
\system{SCDP-2}     & .854 & .692 & .659 & .675 & .227 & .538 & 1.118& .828 \\
\system{SCDP-3}     & .889 & \bf .761 & \bf .751 & \bf .756 & \bf .158 & \bf .710 & \bf 0.791& \bf .878 \\
\hline
\end{tabular}
\caption{Results on the digits data.}
\label{tab:res-digits}
\end{table}

The results of the systems on the digits data are shown in
Table~\ref{tab:res-digits}.  There are several things to note in these
results.  Somewhat surprising is the relatively poor performance of
the \textsc{Binary} model.  Indeed, this model barely does better than
plain K-means, which completely ignores the training data.  Learning a
distance metric, in the \textsc{Xing-K} system, improves results over
standard K-means, and also performs better than the binary classifier.
The ordering of performance of our model, compared to the learned
distance metric, varies by which metric we believe.  According to
\bsf{F-score} and \bsf{NES}, our model is universally better; however,
according to \bsf{VI} and \bsf{NVI}, the distance metric method
outperforms our model 1, but not models 2 and 3.  Finally, on this
data, our untrained model, \textsc{CDP} performs quite poorly, and
makes far too many clusters.

\subsubsection{Identity uncertainty data}

The second data that set we apply our algorithm to is based on the
real world problem of \emph{identity uncertainty} or \emph{entity
integration}.  The data used in the experiment is mined from the dblp
bibliography server.\footnote{Thanks to Anhai Doan, Hui Fang and Rishi
R. Sinha for making this available, see
\url{http://anhai.cs.uiuc.edu/archive/domains/researchers.html} for
further information.}  Each ``publication'' in the data is a computer
science researcher and each ``reference'' is a name occurring in a
reference.  There are a total of 1382 elements in the data,
corresponding to 328 total entities (all labeled).  We use 1004
instances (225 entities) as training data and the rest (378 instances
and 103 entities that do not occur in training) as testing data.

The (pairwise) features we use for this data set are the following:
string edit distance between the two first names; string edit distance
between the two last names; string edit distance between the full
names; Euclidean distance between the publication years; Euclidean
distance between the number of publications (in our data) published in
those years; string edit distance between conference names; Euclidean
distance between the number of publications published in those
conferences; and the number of coauthors with normalized string edit
distance less than $0.1$.  We ran fifty chains of ten thousand
iterations each.  One chain for Models 1 and 2 required approximately
one day to complete, while Model 3 took approximately $3$ days per
chain.

\begin{table}
\center\small
\begin{tabular}{|l||c||c|c|c||c|c||c|c|}
\hline
\makebox[25mm][l]{\bf System} & \makebox[1cm][c]{\bsf{RI}} & \makebox[1cm][c]{\bsf{P}} & \makebox[1cm][c]{\bsf{R}} & \makebox[1cm][c]{\bsf{F}} & \makebox[1cm][c]{\bsf{CED}} & \makebox[1cm][c]{\bsf{NES}} & \makebox[1cm][c]{\bsf{VI}} & \makebox[1cm][c]{\bsf{NVI}} \\
\hline
\system{Coarse}        & .079 & .079 & 1.00 & .147 & .749 & .251 & 3.589 & .395 \\
\system{Fine}          & .921 & \bf 1.00 & .000 & .000 & .000 & .273 & 2.345 & .605 \\
\system{NameMatch}     & .933 & .545 & \bf 1.00 & .706 & .405 & .595 & 1.252 & .789 \\
\hline
\system{K-Means}       & .912 & .451 & .510 & .479 & .341 & .373 & 1.919 & .677 \\
\system{CDP}        & .913 & .480 & .452 & .466 & .355 & .360 & 2.031 & .658 \\
\hline
\system{Binary}        & .855 & .753 & .801 & .776 & .389 & .553 & 1.193 & .808 \\
\system{Xing-K}        & .916 & .467 & .423 & .444 & .378 & .304 & 2.112 & .644 \\
\hline
\system{SCDP-1}     & .963 & .764 & .786 & .775 & .127 & .761 & 0.806 & .864 \\
\system{SCDP-2}     & .971 & .820 & .814 & .817 & .111 & .796 & 0.669 & .887 \\
\system{SCDP-3}     & \bf .982 & \bf .875 & .913 & \bf .894 & \bf .066 & \bf .876 & \bf 0.423 & \bf .929 \\
\hline
\end{tabular}
\caption{Results on the identity uncertainty data.}
\label{tab:res-iu}
\end{table}

We introduce an additional baseline for this data set that groups
person names with identical last names and identical first initials.
This baseline is denoted \system{NameMatch}.  The results of the
systems on the identity uncertainty data are shown in
Table~\ref{tab:res-iu}.  The trend of results here largely agrees with
that of the digits data, in which our models 2 and 3 outperform the
baseline systems.  However, in this case, running the distance-metric
learning algorithm actually hurts the results.  This is perhaps
because our data does not live in Euclidean space, and hence the
optimization performed in learning the distance metric is not run
under the proper conditions.

In this data, according to the \bsf{F-score}, the binary classifier
outperforms our model 1 (though our models 2 and 3 outperform the
binary classifier).  However, according to both the edit distance
metrics and the information metrics, our models all outperform the
binary classifier.  This data also provides a good example of the
deficiencies of the rand index: according to the RI, the \textsc{Fine}
system outperforms all of: \textsc{K-Means, CDP, Binary} and
\textsc{Xing-K}.  Note also in this data that none of the unsupervised
models are able to outperform the \textsc{NameMatch} baseline system
(and neither does the \textsc{Xing-K} system).

\subsubsection{Proper noun coreference data}

The third set of data on which we evaluate is a subtask of the
coreference task, namely, coreference of proper nouns (e.g., ``George
Bush'' $\leftrightarrow$ ``President Bush'' $\leftrightarrow$ ``Bush''
$\not\leftrightarrow$ ``President Clinton'').  This subtask is
significantly simpler than the full task, since one need not identify
coreference between pronouns and proper nouns (``he''
$\leftrightarrow$ ``George Bush''), nor proper nouns and definite
descriptions (``George Bush'' $\leftrightarrow$ ``the President'').
This task has previously been used as a benchmark by
\citet{mccallum04coref}.  We use a partition of the ACE 2004 broadcast
news and newswire training corpus as the training and test data.  This
totals 280 documents for training data and 59 documents for test data.
The training data consists of 5613 mentions of entities, corresponding
to a total of 3100 different entities; the test data contains 950
mentions corresponding to 523 entities.

As features we use string edit distance, string edit distance on the
heads (final word), the length of the longest common substring, the
length of the longest common subsequence, and the string edit distance
between the abbreviations of both terms.  For computing this final
term, we first map words sequences like ``George W. Bush'' to ``GWB''
and leave sequences that already look like abbreviations (eg.,
``IBM'') alone; we then compute string edit distance between these
pairs.  For this data set, we ran fifty chains for ten thousand
iterations.  Models 1 and 2 completed in about three days and Model 3
completed in one week.

\begin{table}
\center\small
\begin{tabular}{|l||c||c|c|c||c|c||c|c|}
\hline
\makebox[25mm][l]{\bf System} & \makebox[1cm][c]{\bsf{RI}} & \makebox[1cm][c]{\bsf{P}} & \makebox[1cm][c]{\bsf{R}} & \makebox[1cm][c]{\bsf{F}} & \makebox[1cm][c]{\bsf{CED}} & \makebox[1cm][c]{\bsf{NES}} & \makebox[1cm][c]{\bsf{VI}} & \makebox[1cm][c]{\bsf{NVI}} \\
\hline
\system{Coarse}              & .003 & .003 & 1.00 & .006 & .978 & .021 & 5.950& .132 \\
\system{Fine}                & .997 & 1.00 & .000 & .000 & 1.00 & .551 & 0.906& .868 \\
\system{SameHead}            & \bf .999 & \bf .965 & \bf .899 & \bf .931 & \bf .019 & \bf .933 & \bf 0.123& \bf .982 \\
\hline
\system{K-Means}             & .994 & .297 & .773 & .429 & .391 & .524 & 1.059& .846 \\
\system{CDP}              & .995 & .273 & .384 & .319 & .352 & .418 & 1.265& .815 \\
\hline
\system{Binary}              & \bf .999 & \bf .900 & \bf .893 & \bf .896 & \bf .040 & \bf .936 & \bf 0.141& \bf .979 \\
\system{Xing-K}              & .998 & .489 & .802 & .608 & .308 & .599 & 0.911& .883 \\
\hline
\system{SCDP-1}           & .996 & .409 & .497 & .449 & .261 & .564 & 0.938& .863 \\
\system{SCDP-2}           & .997 & .596 & .717 & .651 & .203 & .682 & 0.654& .904 \\
\system{SCDP-3}           & \bf .999 & .804 & .882 & .841 & .083 & .861 & 0.284& .958 \\
\hline
\end{tabular}
\caption{Results on the proper noun coreference data.}
\label{tab:res-coref}
\end{table}

As an additional baseline, we cluster mentions with the same head word
(final word); this is denoted \system{SameHead}.  The results of the
systems on the coreference data are shown in
Table~\ref{tab:res-coref}.  As a point of comparison,
\citet{mccallum04coref} report an \bsf{F-score} of $.931$ on this task,
using a graph partition strategy, with weights trained using a
perceptron-style algorithm.  Our binary classification model achieve a
slightly lower \bsf{F-score} or $.896$.  Neither of the unsupervised
algorithms perform very well on this data, but in this data, the
trained distance metric performs better than standard K-means.

Overall the binary classifier is the best of the learned systems,
achieving an F of $.896$, a NES of $.936$ and a normalized variation
of information of $.979$ (compared to our best scores of $.841$,
$.861$ and $.958$, respectively).  However, even the binary classifier
is outperformed along all metrics by the simple baseline that matches
on the final word, \textsc{SameHead}, which achieves scores of $.931$,
$.933$ and $.982$ for the three overall metrics.  The $.931$ \bsf{F-score}
is, incidentally, the same number reported by \citet{mccallum04coref}
(though their choice of training/test division is likely different
from ours).  Overall, however, based on this data, it seems reasonable
to say that one might be better served writing a dozen more rules to
capture notions of abbreviation, post-modification, and a few other
simple phenomena to handle the proper noun coreference task, rather
than try to learn a model from data.\footnote{Of course, the
proper-noun coreference task is the easiest subtask of full
coreference resolution, where empirical results have shown learned
systems are able to outperform rule-based systems.}

\subsubsection{Reference matching data}

Lastly, we perform evaluation on the Cora reference matching data set
\cite{mccallum00matching}.\footnote{Thanks to Andrew McCallum for
making this data available.}  This data consists of all references
from their collection to publications by Michael Kearns, Robert
Schapire and Yoav Freund.  There are 1916 references and 121
publications.  In the original publication, McCallum et al. treated
this as a pure clustering task.  In order to view it as a supervised
clustering task, we treat the labeled data for two of these authors as
training data, using the last author as testing data (performing the
segmentation this way is more realistic than random selection, and
also serves to strengthen the point that the training and testing data
are largely unrelated).

We use the same feature set as in the identity uncertainty evaluation,
with the exception that the first two features become the string edit
distance between the publication names and the string edit distance
between the primary author names, respectively.  Note that this data
is significantly noisier than the data used in the previous section:
there are errors on the labeling of the fields.  We again ran fifty
chains for ten thousand iterations; the chains for Models 1 and 2 took
one day and Model 3 took three days.

\begin{table}
\center\small
\begin{tabular}{|l||c||c|c|c||c|c||c|c|}
\hline
\makebox[25mm][l]{\bf System} & \makebox[1cm][c]{\bsf{RI}} & \makebox[1cm][c]{\bsf{P}} & \makebox[1cm][c]{\bsf{R}} & \makebox[1cm][c]{\bsf{F}} & \makebox[1cm][c]{\bsf{CED}} & \makebox[1cm][c]{\bsf{NES}} & \makebox[1cm][c]{\bsf{VI}} & \makebox[1cm][c]{\bsf{NVI}} \\
\hline
\system{Coarse}              & .118 & .118 & 1.00 & .205 & .745 & .255 & 2.977& .538 \\
\system{Fine}                & .882 & 1.00 & .000 & .000 & .000 & .105 & 3.456& .462 \\
\hline
\system{K-Means}             & .862 & .407 & .461 & .433 & .392 & .240 & 2.655& .577 \\
\system{CDP}              & .850 & .353 & .379 & .365 & .449 & .125 & 2.948& .531 \\
\hline
\system{Binary}              & .936 & \bf .804 & .616 & .686 & \bf .107 & .721 & \bf 0.762& \bf .881 \\
\system{Xing-K}              & .855 & .369 & .384 & .377 & .411 & .180 & 2.807& .552 \\
\hline
\system{SCDP-1}           & .892 & .529 & .507 & .518 & .319 & .372 & 2.237& .643 \\
\system{SCDP-2}           & .934 & .696 & .741 & .718 & .184 & .641 & 1.382& .780 \\
\system{SCDP-3}           & \bf .952 & .794 & \bf .782 & \bf .788 & .125 & \bf .757 & 0.957& .847 \\
\hline
\end{tabular}
\caption{Results on the reference matching data.}
\label{tab:res-rm}
\end{table}

The results of the systems on the reference matching data are shown in
Table~\ref{tab:res-rm}.  In this data, the unsupervised algorithms perform
quite poorly, in comparison to the systems that make use of the training
data.  Again, as in the identity uncertainty data, we see that learning a
distance metric can hurt performance (at least according to \bsf{F-score};
with respect to edit score and normalized \bsf{VI}, it seems to help, but
only marginally so).

According to \bsf{F-score}, the binary classifier on this data outperforms 
our
model 1, though our models 2 and 3 are able to outperform the binary
classifier system.  In terms of edit score, the binary system outperforms
all of our models, except for our model 3, which is able to do slightly
better ($.757$ versus $.721$).  In terms of \bsf{NVI}, the binary
classifier beats all of our models, even model 3, where it achieves an
\bsf{NVI} of $.881$ and we only achieve $.847$.

\subsection{Summary of Results} \label{sec:result-summary}

\begin{table}
\center\small
\begin{tabular}{|l|c|c|c|c|c|}
\hline
& \system{Binary} & \system{Xing-K} & \system{SCDP-1} & \system{SCDP-2} & \system{SCDP-3}\\
\hline
Digits                    & .592 & .608 & .666 & .675 & \bf .756 \\
\hline
Identity Uncertainty      & .776 & .444 & .775 & .817 & \bf .894 \\
\hline
Proper Noun Coreference   & \bf .896 & .608 & .449 & .651 & .841 \\
\hline
Reference Matching        & .686 & .377 & .518 & .718 & \bf .788 \\
\hline
\end{tabular}
\caption{Summary of F-scores of the learning systems on all four data sets.}
\label{tab:res-summary}
\end{table}

We have summarized the results of the five learning systems in
Table~\ref{tab:res-summary} by listing only their final \bsf{F-score}.
Across all data sets, we consistently see that the supervised
approaches outperform the unsupervised approaches, which is not a
terribly surprising finding.  Additionally, the standard K-means
algorithm always outperformed our \textsc{CDP} model (the
unsupervised version of our model).  In two of the data sets (digits
and proper noun coreference), the learned distance metric
(\textsc{Xing-K}) achieved superior performance to standard K-means,
but in the other two data sets, learning the distance metric hurt.  In
those cases, we attribute this loss in performance to the fact that
the algorithm was not operating on truly Euclidean data.

Comparing our models against each other, of our models, model 1 is the
poorest performer, followed by model 2, and model 3 is the best.  Our
models also tend to have higher precision than recall, which suggests
that they create too many clusters.  One could potentially reduce this
by cross-validating on \bsf{F-score} to adjust the $\al^p$ parameter to
attain a balanced precision/recall, but one strong point of Bayesian
models is that no cross-validation is necessary.

Our model 3 was able to outperform the binary classification model in most
metrics on most data sets, but not always.  It tended to consistently
outperform the binary classifier in terms of \bsf{F-score}, but in terms
of \bsf{NES} and \bsf{NVI}, the binary classifier was better on the
reference matching data.  On the proper noun coreference data, our model
was unable to match the performance of the binary classifier, but both
performed more poorly than the simple head-matching baseline system,
suggesting that future work on this subtask is perhaps best handled by
rules, rather than learning.  On the other data sets (digits and identity
uncertainty), our models 2 and 3 consistently outperformed the binary
classification model.

\section{Discussion} \label{sec:discussion}

In this paper, we have presented a Bayesian model for the supervised
clustering problem.  We have dealt with the difficulty of defining a
prior over a potentially infinite set by appealing to the Dirichlet
process prior.  We have introduced the concept of a ``reference type''
as a mechanism for representing the aspects of the data that are
general to the entire data set---essentially allowing for the
supervision.  Like any generative Bayesian classification model, our
framework requires the specification of the data generating
distribution, which we have denoted $F$.  In general, the $F$
distribution is problem-specific, but we have presented a generic
parameterization when $F$ is a Gaussian distribution.

In all but trivial cases, exact evaluation of the posterior
distribution of the class variables in our model is intractable.  We
have presented MCMC-based sampling algorithms that are able to
overcome this intractability.  Unlike deterministic approximation
techniques (such as variational or mean-field inference, or
expectation propagation), the MCMC methods are able to perform even
when non-conjugate priors are employed.  We have presented sampling
algorithms for a full Bayesian treatment of the problem.

Experimentally, under the Gaussian assumption our initial model is
unable to separate classes well.  To fix this problem, we
introduced two subsequent models.  The first modification we make is
to use the references to adjust the parameterization of the prior over
the reference types (model 2).  This enables the use of a sampling
procedure that is essentially as efficient as that used in the
original model (model 1).  The other modification we employ is to
condition the choice of the reference types on the references (model
3).  Unfortunately, in this model, the distributions over the
reference types and the references are no longer conjugate to the data
generating distribution, so a less efficient Gibbs sampler must be
employed to perform inference (in general, a single iteration of the
non-conjugate model is approximately $10$ times slower than one
iteration of the conjugate model).

In a systematic comparison on four data sets against both supervised
and unsupervised models, we have demonstrated that our model is
typically able to attain a higher level of performance than other
models (see Section~\ref{sec:result-summary} for a summary of the
experimental results).  Full Bayesian inference (similar to
transduction) has an advantage over the more standard
training/prediction phases: the test data has no influence on the
reference types.

The largest weakness of our model is its generative nature and the
potential difficulty of specifying a good distribution $F$ that fits
the data and results in tractable inference.  The Gaussian
parameterization seems general, experimentally, but in order to
maintain tractability, we had to assume that the covariance matrix was
diagonal: this is essentially the same as making a na\"ive Bayes
assumption on the features.  For discrete data, using a
multinomial/Dirichlet pair or binomial/beta pair instead of the
normal/gamma pair might be more natural and would lead to nearly the
same inference.  However, like other generative models, it is likely
that our model would be struck with the curse of dimensionality for
any large number of highly correlated features.  The generative story
employed by \citet{pasula03iu} is clearly superior to our---largely
unmotivated---Gaussian assumption; it would be very interesting to
incorporate their generative story into our ``$F$'' distribution,
hopefully to obtain the benefits of both models.

Clearly, scalability is also an issue for our model.  The most
computationally intensive run of our model with the application of
Model 3 to the proper noun coreference data, which required roughly
one CPU \emph{year} to perform.  This is not to say that, for
instance, the binary classification scheme was enormously efficient:
training a cross-validated SVM on this data set took approximately one
CPU month to perform, though this could be improved by not rerunning
the SVM learning for each fold.  Nevertheless, our approach is still
roughly ten times slower.  However, there are several methods that one
can employ to improve the speed of the model, especially if we wish to
scale the model up to larger data sets.  For instance, employing the
canopy method described by \citet{mccallum00matching} and only
considering drawing the $c$ indicator variables from appropriate
canopies would drastically improve the efficiency of our model,
provided the canopies were sufficiently small.  Furthermore, in the
cases of Models 1 and 2, since conjugate priors \emph{are} used, one
could employ a more efficient sampling scheme, similar to the
Metropolis-Hastings algorithm suggested by \citet{xing04haplotype} or
the split-merge proposals suggested by \citet{jain-neal03split-merge}.
Nevertheless, MCMC algorithms are notoriously slow and experiments
employing variational or EP methods for the conjugate models might
also improve performance \citep{blei05variational,minka04dpep}.

Our model is also similar to a distance metric learning algorithm.
Under the Gaussian assumption, the reference types become covariance
matrices, which---when there is only one reference type---can be
interpreted as a transform on the data.  However, when there is more
than one reference type, or in the case of full Bayesian inference,
the sorts of data distributions accounted for by our model are more
general than in the standard metric learning
scenario.\footnote{Consider, for instance, a two dimensional Euclidean
space where the clusters are axis-aligned pluses.  Our model learns
two ``reference types'' for this data: one aligned with each axis,
and, for data that is reasonably separated, is able to correctly
classify most test data.  On the other hand, a metric learning
algorithm cannot perform any linear transformation on the data that
will result in ``better looking'' clusters.}

We believe future research in the context of the framework described
in this paper can proceed along several dimensions.  The most obvious
would be the integration of more domain-specific information in the
data generating distribution $F$.  One might be also able to achieve a
similar effect by investigating the interaction of our model with
various unsupervised embedding techniques (kPCA, LLE, MDS, etc.).  We
have performed preliminary investigations using kPCA (using the
standard string kernel) and LLE combined with K-means as well as
K-means and distance-metric learning and have found that performance
is substantially worse than the results presented in this paper.  A
final potential avenue for future work would be to attempt to combine
the power of our model with the ability to incorporate arbitrary
features found in conditional models, like that of
\citet{mccallum04coref}.  Such an integration would be technically
challenging, but would likely result in a more appropriate, general
model.

Finally, to foster further research in the supervised clustering
problem, we have contributed our data sets and scoring software to the
RIDDLE data repository,
\url{http://www.cs.utexas.edu/users/ml/riddle/}, maintained by Mikhail
Bilenko.


\acks{The authors would like to thank Aaron D'Souza for several
  helpful discussions of the Dirichlet process.  We would also like to
  thank the anonymous reviewers of an earlier draft of this paper,
  whose advice improved this paper dramatically, as well as the three
  anonymous reviewers of the current version of this paper who
  contributed greatly to its clarity and content.  Some of the
  computations described in this work were made possible by the High
  Performance Computing Center at the University of Southern
  California.  This work was partially supported by DARPA-ITO grant
  N66001-00-1-9814, NSF grant IIS-0097846, and a USC Dean Fellowship
  to Hal Daum\'e III. }

\appendix
\section*{Appendix A. The Dirichlet Process}

The formal definition of the Dirichlet process is as follows.  Let
$(\cX, \Om)$ be a measurable space and let $\mu$ be a measure
(unnormalized density) on this space that is finite, additive,
non-negative and non-null.  We say that a random probability measure
$P^\mu$ on $(\cX, \Om)$ is a \emph{Dirichlet process} with parameter
$\mu$ under the following condition: whenever $\{ B_1, \dots, B_K \}$
is a measurable partition of $\Om$ (i.e., each $\mu(B_k)>0$ for all
$k$) , then the joint distribution of random probabilities
$(P^\mu(B_1), \dots, P^\mu(B_K))$ is distributed according to
$\Dir(\mu(B_1), \dots, \mu(B_K))$, where $\Dir$ denotes the standard
Dirichlet distribution
\citep{ferguson73bayesian,ferguson74priordistributions}.  In words:
$P^\mu$ is a Dirichlet process if it behaves as if it were a Dirichlet
distribution on any finite partition of the original space.

It is typically useful to write $\mu = \al G_0$, where $\al = \int_\Om
\ud \mu$ and $G_0 = \mu / \al$, so that $G_0$ is a density.  In this
case we refer to $G_0$ as the \emph{base distribution} or the
\emph{mean distribution} of the DP, and $\al$ as the \emph{precision},
or \emph{scale parameter}.

Two fundamental results regarding the DP that are important to us are:
(1) observations from a DP are discrete (with probability one) and (2)
if $P^\mu$ is a DP with parameter $\mu$, then the conditional
distribution of $P^\mu$ given a sample $X_1, \dots, X_N$ is a DP with
parameter $P^\mu + \sum_{n=1}^N \de_{X_n}$, where $\de_X$ is a point
mass concentrated at $X$ \citep{ferguson74priordistributions}.  The
final useful fact is a correspondence between the DP and P\`olya Urns,
described by \citet{blackwell-macqueen73polya}.  In the P\`olya Urn
construction, we consider the situation of an urn from which we draw
balls.  Initially the urn contains a single black ball.  At any time
step, we draw a ball $x$ from the urn.  If $x$ is black (as it must be
on the first draw), we put $x$ back into the urn and also add a ball
of a brand new color.  If $x$ was not black, we put $x$ back into the
urn and also put in an additional ball of the same color.  The pattern
of draws from such an urn describes draws from a DP (with $\al=1$).
In this scheme, we can see that there is a clustering effect in this
model: as more balls of one color (say, blue) are drawn, the number of
blue balls in the urn increases, so the probability of drawing a blue
ball in the next iteration is higher.  However, regardless of how many
balls there are in the urn, there is always some probability the black
ball (i.e., a ball of a new color) is drawn.  This relative
probability is controlled by the precision parameter $\al$.  For low
$\al$, there will be few colors and for high $\al$, there will be many
colors.  The appropriateness of such a prior depends on one's prior
intuitions about the problem; more flexible similar priors are given
in terms of exchangeable probability partition functions, including a
simple two-parameter extension of the DP, by \citet{pitman96urn}.

As noted by \citet{ferguson83mixtures}, the discreteness of
observations from the DP means that observations from the
distributions drawn from a DP can be viewed as countably infinite
mixtures.  This can be seen directly by considering a model that first
draws a distribution $G$ from a DP with parameter $\al G_0$ and then
draws observations $\th_1, \dots$ from $G$.  In such a model, one can
analytically integrate out $G$ to obtain the following conditional
distributions from the observations $\th_n$
\citep{blackwell-macqueen73polya,ferguson83mixtures}:

\begin{equation*} 
\th_{n+1} \| \th_1, \dots, \th_n
        ~\by~ \frac \al {n+\al} ~ G_0 + \frac 1 {n+\al} \sum_{i=1}^n \de_{\th_i}
\end{equation*}

Thus, the $n+1$st data point is drawn with probability proportional to
$\al$ from the base distribution $G_0$, and is exactly equal to a
previously drawn $\th_i$ with probability proportional to
$\sum_{j=1}^n \de_{\th_i = \th_j}$.  This characterization leads to a
straightforward implementation of a Gibbs sampler.  It also enables
one to show that the posterior density of a DP with parameter $\mu$
after observing $N$ observations $\th_1, \dots, \th_N$ is again a DP
with parameter $\mu + \sum_{n=1}^N \de_{\th_n}$
\citep{ferguson73bayesian}.

\section*{Appendix B. Sampling the Precision Parameter}

\citet{west92hyperparam} describes a method of sampling the precision
parameter $\al$ for a DP mixture model.  Placing a $\Gam(a,b)$ prior
over $\al$, when $n$ (the number of observations) and $k$ (the number
of unique mixture components) are known, one first samples an
intermediary value $x$ by a beta distribution $x^\al(1-x)^{n-1}$,
where $\al$ is the previous value for the precision parameter.  Given
this random variable $x$, one resamples $\al$ according to a mixture
of two Gamma densities:

\begin{equation*}
\pi_x \Gam(a+k, b-\log x) + (1-\pi_x) \Gam(a+k-1, b-\log x)
\end{equation*}

Where $\pi_x$ is the solution to $\pi_x / (1-\pi_x) = (a+k-1) /
[n(b-\log x)]$.  To extend this method to the case with multiple $n$
and $k$, we first recall the result of \citet{antoniak74mixtures},
which states that the prior distribution of $k$ given $\al$ and $n$ is
given by:

\begin{equation*}
\p{{k} \| \al, {n}} = c_{n}({k}){n}!\al^{k}\frac {\Ga(\al)} {\Ga(\al+{n})}
\end{equation*}

Here, $c_n(k) \propto |S_n^{(k)}|$, a Stirling number of the first
kind, does not depend on $\al$.  Placing a gamma prior on $\al$ with
shape parameter $a$ and scale parameter $b$, we obtain the posterior
distribution of $\al$ given all the $n_m, k_m$ as:

\begin{eqnarray}
\p{\al \| \vec x, \vec k, \vec n} 
&\propto&
  e^{-b\al} \al^{a-1}
  \prod_{m=1}^M \al^{k_m-1} (\al+n_m) x_m^\al(1-x_m)^{n_m-1}
  \nonumber \\
&\propto&
  \al^{a-M-1+\sum_{m=1}^M k_m} e^{-\al\(b-\log \prod_{m=1}^M x_m\)}
  \prod_{m=1}^M (\al+n_m)
  \label{eq:marginal-al2}
\end{eqnarray}

The product in Eq~\eqref{eq:marginal-al2} can be written as the sum over
a vector of binary indicator variables $\vec i$  of length $M$, which
gives us:

\begin{equation} \label{eq:al-mixture-marginal}
\al \| \vec x, \vec k, \vec n \by
  \sum_{\vec i \in 2^M}
    \rh_{\vec i} \Gam\(a-M+\sum_{m=1}^M k_m + i_m, b - \log \prod_{m=1}^M x_m\)
\end{equation}

Where, writing $\hat a$ to denote the value $a - M - 1 + \sum_{m=1}^M
k_m$ and $\hat b$ to denote $b - \log \prod_{m=1}^M x_m$, the mixing
weights $\vec \rh$ are defined by:

\begin{equation} 
\rh_{\vec i} = \frac 1 Z
  \Ga\( \hat a + \sum_{m=1}^M i_m \)
  \prod_{m=1}^M 
    \( n_m \hat b \) ^{1-i_m}
  \label{eq:weights}
\end{equation}

To see the correctness of this derivation, consider a given $\vec i$.
There are $\sum i_m$ choices of $\al$, corresponding to the $\sum i_m$
in the shape parameter for the posterior gamma distribution in
Eq~\eqref{eq:al-mixture-marginal}.  For each of these, the constant
from the gamma distribution is decreased by a factor of $\Ga(\hat a +
\sum i_m) / \Ga(\hat a)$; compensating for this results in the first
term above (with the bottom half omitted since it is just a constant).
Additionally, each term for which $i_m=0$ means that $n_m$ was chosen
(instead of $\al$), so a factor of $n_m = n_m^{1-i_m}$ needs to be
included.  Finally, when the shape parameter of the gamma distribution
increases by $1$ for each $i_m=1$, the constant of proportionality for
the gamma distribution increases by a factor of $b - \log \prod x_m$,
which is compensated for by the last term above.

Similarly, we can obtain a marginal distribution for each $x_m$
conditional on $\al$ and $k$ as:

\begin{equation}
x_m \| \al, n_m, k_m ~~\propto~~ x_m^\al (1-x_m)^{n_m-1}
~~\by~~ \Bet(\al+1,n_m) \label{eq:post-xm}
\end{equation}

In order to sample $\al$, we first sample $\vec x$ by a sequence of
$m$ beta distributions according to Eq~\eqref{eq:post-xm}, conditioned
on the current value of $\al$ and $\vec n$.  Then, given these values
of $\vec x$, we sample a new value of $\al$ from a mixture of gammas
defined in Eq~\eqref{eq:al-mixture-marginal}, conditional on the newly
sampled $\vec x$, with weights defined in Eq~\eqref{eq:weights}.  In
the latter step, we simply select an $\vec i \in 2^M$ according to the
probability density $\rh_{\vec i}$ and then sample a value from the
corresponding gamma distribution.

Unfortunately, in all but trivial cases, $M$ is large and so
computing $\rh_{\vec i}$ directly for all such $\vec i$ requires
an exponential amount of time (in $M$).  Thus, instead of computing
the $\vec \rh$s directly, we sample for them, effectively
computing the constant $Z$ though standard MCMC techniques.  To
perform the actual sampling from $2^M$, we employ a Gibbs sampler.
Each iteration of the Gibbs sampler cycles through each of the $M$
values of $\vec i$ and replaces $i_m$ with a new value, sampled
according to its posterior, conditional on $\vec i_{-m} = \langle i_l
\| 1 \leq l \leq M,~ l \ne m \rangle$.  The derivation of this
posterior is straightforward:

\begin{eqnarray}
i_m=1 \| \vec i_{-m}
&=& \frac
      {\hat a + \sum_{m' \ne m} i_m}
      {\hat a + \sum_{m' \ne m} i_m + n_m\hat b}
\label{eq:sample-im}
\end{eqnarray}

Putting it all together, we sample a new value of $\al$ by first
sampling a vector $\vec x$, where each $x_m$ is sampled according to
Eq~\eqref{eq:post-xm}.  Then, we sample $R$-many $\vec i^{(r)}$s using
the Gibbs sampler with update given by Eq~\eqref{eq:sample-im};
finally selecting one of the $\vec i^{(r)}$ according to its empirical
density.  Finally, given this $\vec i$ and the $x_m$s, we sample a new
value for $\al$ by a gamma distribution according to
Eq~\eqref{eq:al-mixture-marginal}.  We have found that for modest $M <
100$, $n_m < 1000$ and $k_m < 500$, such a chain converges in roughly
50 iterations.  In practice, we run it for 200 iterations to be safe.

\vskip 0.2in
\bibliography{bibfile}

\end{document}